\pdfoutput=1

\documentclass[11pt]{article}

\usepackage[final]{acl}

\usepackage{times}
\usepackage{latexsym}
\usepackage{amsmath}
\usepackage{amsfonts}
\usepackage[most]{tcolorbox} 
\tcbuselibrary{breakable,listings}

\usepackage{wrapfig}
\usepackage{caption, subcaption}
\usepackage{listings}

\usepackage{tkz-euclide}
\usepackage{threeparttable}
\usepackage{algpseudocode}
\usepackage{flushend}
\usepackage{booktabs}                   
\usepackage{multirow}
\usepackage{tikz}                       
\usepackage{amssymb}
\usepackage{tcolorbox}
\tcbuselibrary{theorems}                 
\newtcbtheorem[number within=section]{prompt}{Input}%
{colback=green!1,colframe=green!25!black,fonttitle=\bfseries}{th}

\usepackage{array}

\newcommand{\ntpp}{\mathsf{ntpp}}
\newcommand{\tpp}{\mathsf{tpp}}
\newcommand{\ntppi}{\mathsf{ntppi}}
\newcommand{\tppi}{\mathsf{tppi}}
\newcommand{\ec}{\mathsf{ec}}
\newcommand{\dc}{\mathsf{dc}}
\newcommand{\eq}{\mathsf{eq}}
\newcommand{\po}{\mathsf{po}}

\newcommand{\intervald}{\mathsf{d}}
\newcommand{\intervaldi}{\mathsf{di}}
\newcommand{\intervalo}{\mathsf{o}}
\newcommand{\intervaloi}{\mathsf{oi}}
\newcommand{\intervalm}{\mathsf{m}}
\newcommand{\intervalmi}{\mathsf{mi}}
\newcommand{\intervals}{\mathsf{s}}
\newcommand{\intervalsi}{\mathsf{si}}
\newcommand{\intervalf}{\mathsf{f}}
\newcommand{\intervalfi}{\mathsf{fi}}

\newtcolorbox{cotbox}[1][]{%
  enhanced,
  breakable,            
  colframe=green!25!black,       
  colback=green!1,
  title={Chain-of-Thought},
  attach title to upper,      
  listing engine=listings,
  listing only,
  #1                         
}
\usepackage[T1]{fontenc}

\usepackage[utf8]{inputenc}

\usepackage{microtype}

\usepackage{inconsolata}

\usepackage{graphicx}

\title{Large Language and Reasoning Models are Shallow Disjunctive Reasoners}

\author{Irtaza Khalid${^1}$, Amir Masoud Nourollah${^1}$, Steven Schockaert${^1}$ \\
         ${^1}$School of Computer Science and Informatics, Cardiff University, United Kingdom
         \\ \texttt{\{khalidmi,nourollaha,schockaerts1\}@cardiff.ac.uk}}

\begin{document}
\maketitle

\begin{abstract}
Large Language Models (LLMs) have been found to struggle with systematic reasoning. Even on tasks where they appear to perform well, their performance often depends on shortcuts, rather than on genuine reasoning abilities, leading them to collapse on out-of-distribution (OOD) examples. Post-training strategies based on reinforcement learning and chain-of-thought prompting have recently been hailed as a step change. However, little is known about the potential of the resulting ``Large Reasoning Models'' (LRMs) beyond maths and programming-based problem solving, where genuine OOD problems can be sparse. In this paper, we focus on tasks that require systematic relational composition for qualitative spatial and temporal reasoning. The setting allows fine control over problem difficulty to precisely measure OOD generalization. We find that, zero-shot LRMs generally outperform their LLM counterparts in single-path reasoning tasks but struggle in the multi-path setting. Whilst showing comparatively better results, fine-tuned LLMs are also not capable of multi-path generalization. We also provide evidence for the behavioral interpretation for this, i.e., that LRMs are shallow disjunctive reasoners. 
\end{abstract}

\section{Introduction}
Large Language Models (LLMs) have shown remarkable generalization abilities, being able to learn from in-context demonstrations, and to generalize to unseen tasks in multi-task settings \cite{radford2019language, brown2020language, bubeck2023sparks}, with abilities in mathematics and programming that appear to go beyond the level of high-school students \cite{guo2024deepseek, jimenez2024swebench, openai2025competitiveprogramminglargereasoning}. 
Moreover, recent advances in post-training based on reinforcement learning have unlocked a further axis along which the ability of LLMs can be improved, for easily verifiable analytical problems (such as mathematics and programming) 
\cite{guo2025deepseek, shao2024deepseekmath, sprague2024cotcotchainofthoughthelps}. The resulting models, called Large Reasoning Models (LRMs), are then encouraged to leverage chains-of-thought (CoT) or thinking tokens \cite{DBLP:conf/nips/Wei0SBIXCLZ22} to search though a solution space, which provably increases the complexity of problems that can be tackled \cite{feng2023towards}, compared to standard LLM prompting.      

Yet, a competing narrative is that current LLMs are not, in fact, general-purpose reasoners and rather rely on shallow pattern matching \cite{dziri2023faith, mccoy2024embers,nguyen2024understanding} and heuristics \cite{DBLP:journals/corr/abs-2410-21272}. There are recurring issues, even with the latest LLMs and LRMs, such as memorization of training data \cite{zhang2024a}, the reversal curse \cite{berglund2024the} and an over-reliance on co-occurrence statistics \cite{kang-choi-2023-impact}. This line of argument is further bolstered by the risk that popular static benchmarks, such as GSM8k and MMLU, may have been included in training corpora \cite{zhang2024a, oren2024proving}. The potential for dataset contamination is increasingly problematic, given the scaling laws for memorization \cite{carlini2023quantifying}, and may explain why despite displaying erudite behaviour, current models still fail at seemingly basic tasks that are trivial for ordinary humans.

In this paper, we highlight the importance of using benchmarks that require Systematic Generalization (SG) for reliably evaluating the reasoning capabilities of LLMs and LRMs. SG is the ability of a model to solve test instances by composing knowledge that was learned from multiple training instances \cite{hupkes2020compositionality}, where the test instances are  systematically made larger than the informationally complete training instances. 
Composing atomic units into larger pieces for constructing a solution to an arbitrarily large problem is an essential ingredient for machines and humans to generalize from a limited amount of data \cite{lake2017building}. 
We specifically advocate the use of synthetic benchmarks, where the difficulty of problem instances can be controlled along different dimensions. 

For the analysis in this paper, we leverage the Spatial Temporal and Reasoning (STaR) benchmark \cite{khalid2025systematic}. 
Its problem instances have a combinatorial structure, which makes it straightforward to generate large numbers of previously unseen cases, and in particular avoid issues of dataset contamination. The StaR benchmark has proven challenging for state-of-the-art neuro-symbolic reasoning methods \cite{ctp, ncrl, r5}, but has not yet been used for evaluating LLMs and LRMs.
%
%
It poses an interesting challenge, because the disjunctive nature of the rules that govern the reasoning problems means that the answer cannot be obtained by a single derivation (i.e.\ a single chain-of-thought) and essentially requires simulating the algebraic closure algorithm \cite{algebraic-closure}. Note, however, that these problems are computationally tractable (i.e.\ they can be solved in polynomial time) and should thus, in principle, be within the reach of LRMs. This is fundamentally different from evaluating LRMs on PSPACE-hard planning problems, where at best strong heuristic approximations can be expected \cite{DBLP:journals/corr/abs-2409-13373}.

\begin{figure*}[t!]
    \centering 
\vspace{10ex}
  \centerline{\includegraphics[width=0.9\linewidth]{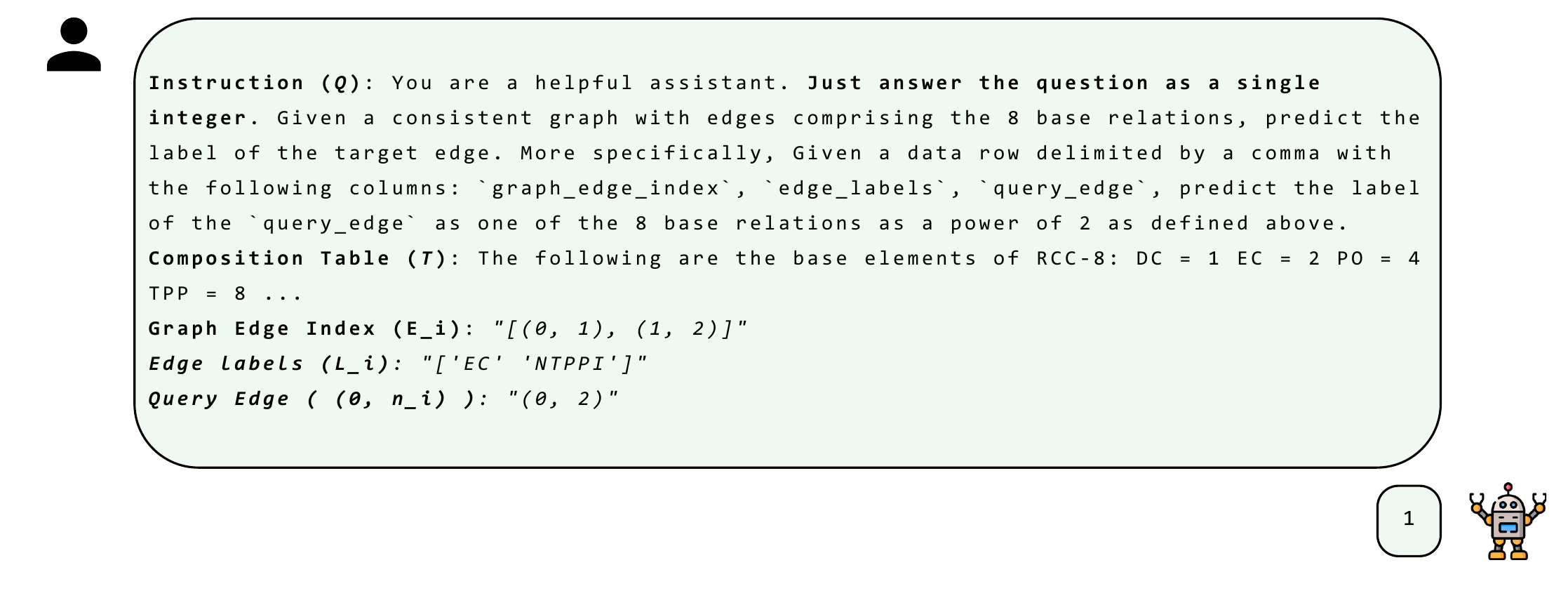}}
\caption{An illustration of the input representation to the language model which is prompted to respond (modulo thinking tokens) with a single label for the query edge.}
\label{fig:illustration-figPrompt}
\end{figure*}
\begin{figure}[t!]
    \centering 
    \includegraphics[width=220pt]{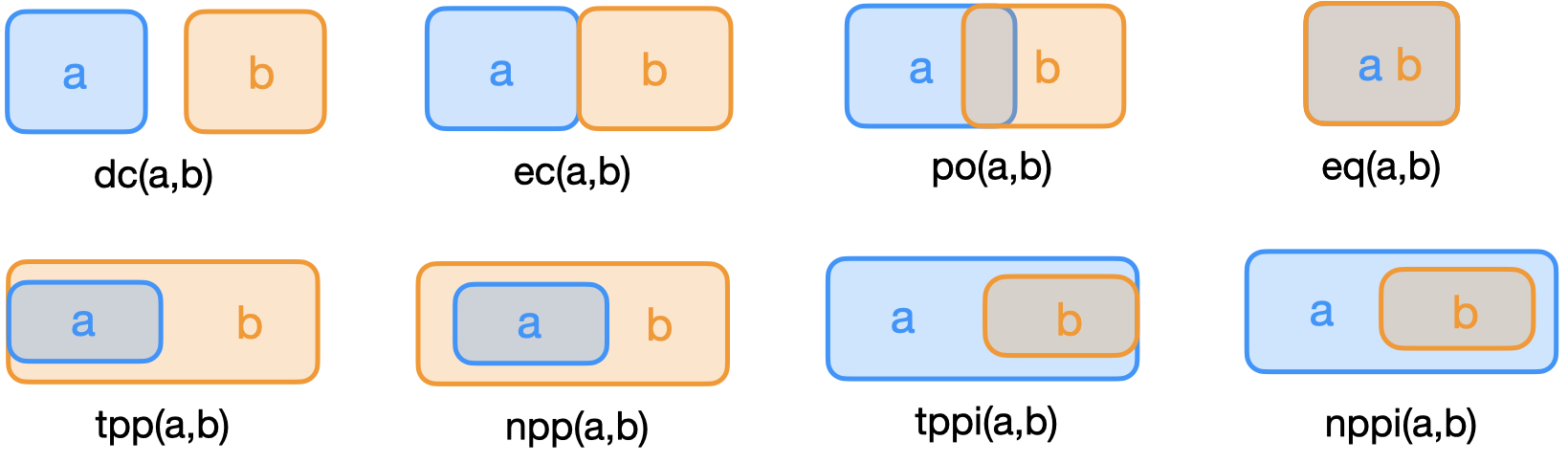}
\caption{Illustration of the RCC-8 relations. }
\label{fig:illustration-figRCC8}
\end{figure}

Our main finding is that many popular LLMs and LRMs struggle on STaR but do reason beyond random chance. We find that LRMs are remarkably able to zero-shot exploit a trivial path heuristic. We analyze the effects of increasing model size, fine-tuning and CoT test-time compute on reasoning performance and provide a behavioral interpretation behind the reasoning abilities of models showing that they shallowly simulate the algebraic closure algorithm required for disjunctive reasoning.

\section{Related Work}

\paragraph{Spatial Reasoning}
The spatial reasoning capabilities of LLMs have already been studied from various angles. For instance, SPARTQA \cite{mirzaee-etal-2021-spartqa}, StepGame \cite{DBLP:conf/aaai/ShiZL22} and RoomSpace \cite{DBLP:conf/ijcai/LiH024} are question answering datasets which require the model to infer the relative position of two objects based on a description of their position relative to other objects. 
However, rather than focusing on qualitative reasoning, these tasks involve geometric computations, e.g.\ determining if a given point belongs to some region or determining the regions through which a given trajectory passes. \citet{DBLP:journals/corr/abs-2411-19589} evaluate whether LLMs can infer the composition of two RCC-8 relations, when given a description of their meaning, while \citet{DBLP:conf/cosit/0001B24} evaluate their commonsense understanding of cardinal directions
. \citet{DBLP:journals/corr/abs-2406-14852} consider spatial reasoning in a multi-modal setting. 

Several authors have also tried to improve LLM spatial reasoning. \citet{DBLP:conf/aaai/LiH024a} study the effectiveness of chain-of-thought \cite{DBLP:conf/nips/Wei0SBIXCLZ22} and tree-of-thoughts \cite{DBLP:conf/nips/YaoYZS00N23} prompting. They also show the effectiveness of using the LLM for semantic parsing and leaving the reasoning to a symbolic solver. 
\citet{DBLP:journals/corr/abs-2404-03622} improve chain-of-thought methods for spatial reasoning, by generating a visualization after each inference step. In multimodal settings, pre-training on synthetic data is common. Interestingly, \citet{DBLP:journals/corr/abs-2410-16162} found that by training the model on basic (visual) spatial reasoning capabilities
, the model also performs better on out-of-distribution composite tasks, such as finding the shortest path between two objects.

\paragraph{Systematic Generalization}
There is a plethora of work on measuring Systematic Generalization (SG) beyond relational reasoning, including SCAN \cite{lakeandbaroni2018} for RNNs \cite{rnn}, addition \cite{addition} and LEGO \cite{LEGO}, for trainable transformers \cite{vaswani2017attention}. These works suggests that transformers struggle with SG. The most popular benchmark for SG for relational reasoning is CLUTRR \citet{clutrr}, which involves predicting family relations. 
\citet{zhu2024largelanguagemodelslearn} evaluated LLMs on this benchmark, showing that even modern LLMs with CoT prompting struggle with this task. The problems we consider in this paper are more challenging than those in CLUTRR, due to the need for combining multiple reasoning paths. 
\paragraph{Rule-based Reasoning with LLMs}
\citet{DBLP:journals/corr/abs-2407-08440} studied the ability of LLMs to apply a given rule, when provided as part of the prompt. In contrast to our experiments in this paper, they only evaluated the application of a single rule, some of which were complex (e.g.\ encoding the composition of a path of several relations). They found chain-of-thought prompting to be largely ineffective, which appears to be related to the fact that multi-hop reasoning was not required for their benchmark. 
They also found evidence that models rely on prior knowledge about the considered domains (e.g.\ the composition of family relations).

\section{The STaR Problem}
In each problem instance of STaR, we are given a set of facts $\mathcal{F}$, referring to a set of binary relations $\mathcal{R}$ and a set of entities $\mathcal{E}$. The set of relations is fixed across problem instances, but the entities are not. Each of the facts is an \emph{atom} of the form $r(a,b)$, with $r\in\mathcal{R}$ and $a,b\in\mathcal{E}$. The problems we consider essentially require models to learn a set of rules $\mathcal{K}$, which they can then use to decide whether a given atom $r(a,b)$ can be inferred from the set of facts $\mathcal{F}$. To be successful, models must be capable of composing the learned rules in a systematic way. In particular, most problem instances require multiple rule applications to be chained, and the number of such inference steps may be larger for test examples than for training examples.

\paragraph{Disjunctive Rules}
Most reasoning benchmarks focus on Horn rules of the following form ($k\geq 3$):
\begin{align}\label{eqRuleWithNAtoms}
r(X_1,X_k) \gets  \bigwedge_{i=1}^{k-1} r_{i}(X_{i},X_{i+1})
\end{align}
where $X_i$ are entity variables. Given a set $\mathcal{K}$ of such rules, the main reasoning task of interest is typically to decide whether some hypothesis $r_\ell(e,f)$ can be inferred from a set of facts $\mathcal{F}$ using the rules in $\mathcal{K}$. This can be decided by repeatedly selecting facts $r_1(e_1,e_2),...,r_{k-1}(e_{k-1},e_k)$ that match the body of a rule of the form \eqref{eqRuleWithNAtoms} in $\mathcal{F}$ and adding the conclusion $r(e_1,e_k)$ of that rule to $\mathcal{F}$. This iterative derivation of facts is well-aligned with the style of reasoning that is enabled by chain-of-thought prompting, which can partially explain the success of such strategies for tasks that require simple logical reasoning.
However, in many domains, Horn rules are not sufficient for capturing the required knowledge. A more general approach is to focus on disjunctive rules of the following form:
\begin{align}\label{eqDisjunctiveRule}
\bigvee_{i=1}^m s_l(X_1,X_k) \gets \bigwedge_{i=1}^{k-1} r_{i}(X_{i},X_{i+1})
\end{align}
Given such a rule and the facts $r_1(e_1,e_2),...,r_{k-1}(e_{k-1},e_k)$, then all we can infer is that one of $s_1(e_1,e_k),...,s_m(e_1,e_k)$ must be true. When reasoning with disjunctive rules, we are typically also given a set of constraints, such as:
$$
\bot \gets r_1(X,Y) \wedge r_2(X,Y)
$$
encoding that at most one of the facts $r_1(e,f), r_2(e,f)$ can be true for any entities $e,f$. Reasoning with disjunctive rules is provably more expressive, but computationally also more expensive: while reasoning with Horn rules is possible in polynomial time, reasoning with disjunctive rules and constraints is an NP-complete problem. However, there are important special cases where reasoning with disjunctive rules is still possible in polynomial time. This is the case, in particular, for many of the calculi that have been proposed for qualitative reasoning about time and space, such as the Interval Algebra (IA \cite{interval}) and the Region Connection Calculus (RCC8 \cite{rcc8}).\footnote{When the set $\mathcal{F}$ is allowed to contain disjunctions of facts, then reasoning with these calculi is NP-complete. However, since we only focus on the case where $\mathcal{F}$ is a set of facts, reasoning for our purposes is tractable in these calculi.}

\paragraph{StaR Benchmark} 
STaR \cite{khalid2025systematic} consists of spatial and temporal reasoning problems. The spatial reasoning problems involve reasoning in RCC-8 \cite{rcc8}. This calculus is defined using 8 relations, illustrated in Fig.~\ref{fig:illustration-figRCC8}. The entities in this case represent spatial regions. For instance, the fact $\mathsf{ec}(a,b)$ specifies that the region $a$ is adjacent (i.e.\ Externally Connected) to the region $b$. Reasoning in RCC-8 is based on two types of knowledge. First, we have the knowledge that the relations are Jointly Exhaustive and Pairwise Disjoint (JEPD), meaning that there is exactly one of the eight relations that holds between any two regions. Second, we have knowledge about the composition of the eight relations. For instance, knowing that $\mathsf{ec}(a,b)$ and $\mathsf{po}(b,c)$ hold, the relations that may hold between $a$ and $c$ are $\mathsf{dc}$, $\mathsf{ec}$, $\mathsf{po}$, $\mathsf{tpp}$ and $\mathsf{ntpp}$.  This knowledge can be encoded using a disjunctive rule. It is typically summarized in a so-called composition table, which encodes the compositions of all relations in a compact format. The temporal instances in StaR involve reasoning in IA \cite{interval}. The overall structure of these reasoning problems is similar as in RCC-8, but here there is a set of 13 JEPD relations. The entities in this case represent time intervals, and we have relations such as $\mathsf{m}(e,f)$, encoding that the end point of $e$ coincides with the starting point of $f$.

 Each problem instance is formulated as a directed labelled graph $\mathcal{G}$, where the vertices represent entities and the edges are labelled with a relation from $\mathcal{R}$, where $\mathcal{R}$ is either the set of RCC-8 relations or the set of IA relations. The goal is to infer the relationship that holds between two designated entities: a source entity $s$ and a tail entity $t$. The problem instances are constructed such that there is a unique relation that can be inferred. To find this relation, however, information from multiple paths between $s$ and $t$ may need to be combined. Each of these paths makes it possible to infer a conclusion of the form $r_1(s,t)\vee ... \vee r_m(s,t)$. In other words, each path allows us to eliminate certain relationships as candidate answers, but we may need to combine several paths to eliminate all-but-one of the relations and thus obtain the answer.
 The dataset is constructed with two levers of complexity: $b$, the number of simple paths between the source and tail entity, and $k$, the length of each simple path. 
 In accordance with the focus on SG, the training or fine-tuning data is comprised of small problem instances, with $k\in\{2,3,4\}$ and $b\in\{1,2,3\}$. The test data contains instances with $k \in \{2, \dots, 10\}$ and $b \in \{1,2,3,4\}$.
 
\begin{table}[t!]
\footnotesize   
\setlength\tabcolsep{3pt}
    \centering    
        \begin{tabular}{clccccc} 
            \toprule
            & \textbf{Model} & \textbf{Param.} & \multicolumn{3}{c}{\textbf{Quantization}} & \textbf{Reasoning} \\
             \cmidrule(lr){4-6} 
            & &  & A & B & C &  \\ 
            \midrule
            \parbox[t]{3mm}{\multirow{4}{*}{\rotatebox[origin=c]{90}{\textbf{Small}}}}
            & Qwen-2.5                & 7B  & \texttimes & \texttimes & \tikz\node{\checkmark}; & N/A  \\
            & Qwen-2.5 \textbf{(R)}   & 7B & \texttimes & \texttimes & \tikz\node{\checkmark}; & \tikz\node{\checkmark};  \\
            & Llama-3                 & 8B  & \texttimes & \texttimes & \tikz\node{\checkmark}; & N/A  \\
            & Gemma-2                 & 9B  & \texttimes & \texttimes & \tikz\node{\checkmark}; & N/A  \\
            \midrule
             \parbox[t]{3mm}{\multirow{4}{*}{\rotatebox[origin=c]{90}{\textbf{Medium}}}} &
            Phi-4                     & 14B  & \texttimes & \texttimes & \tikz\node{\checkmark}; & N/A  \\
            & Qwen-2.5                & 14B       & \texttimes & \texttimes & \tikz\node{\checkmark}; & N/A  \\
            & Qwen-2.5 \textbf{(R)}   & 14B  & \texttimes & \texttimes & \tikz\node{\checkmark}; & \tikz\node{\checkmark};  \\
            & Gemma-2                 & 27B  & \texttimes & \texttimes & \tikz\node{\checkmark}; & N/A  \\
            \midrule
             \parbox[t]{3mm}{\multirow{4}{*}{\rotatebox[origin=c]{90}{\textbf{Large}}}} &
            Llama-3.3                 & 70B  & \tikz\node{\checkmark}; & \tikz\node{\checkmark}; & N/A & N/A  \\
            & Qwen-2.5                & 72B  & \tikz\node{\checkmark}; & \tikz\node{\checkmark}; & N/A & N/A  \\
            & o3-mini                 & ?  & N/A & N/A & N/A & \tikz\node{\checkmark}; \\
            \bottomrule
        \end{tabular}
        \caption{Model configurations for experimental settings in \ref{sec:experiments}. All the quantizations are four-bit. \textbf{(R)} denotes the R1 distilled models \cite{guo2024deepseek}. }\label{table:model-stats}
\end{table}

\begin{figure*}[t!]
    \centering
    \includegraphics[width=0.95\linewidth]{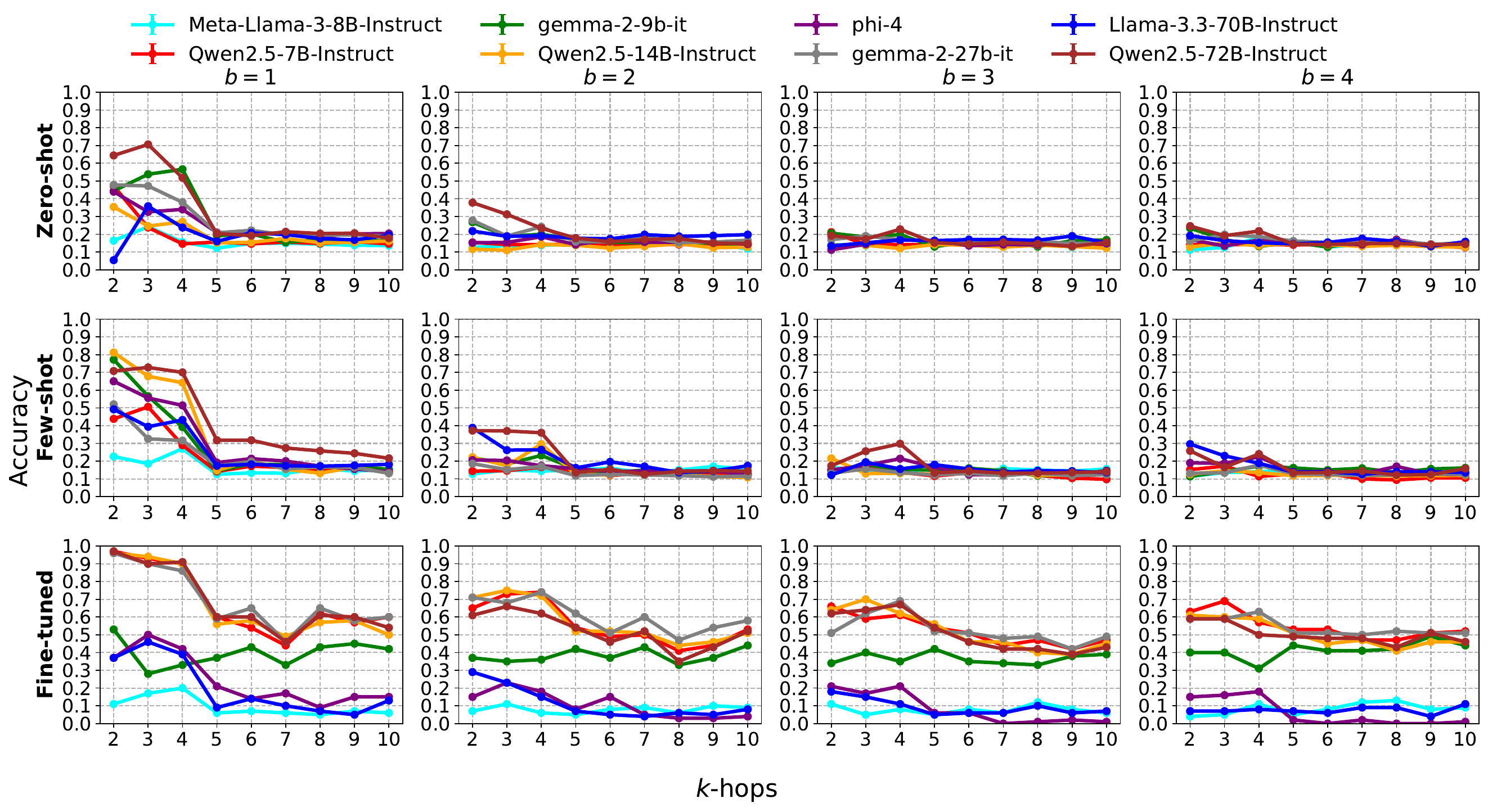}
    \caption{The results for the non-reasoning models on RCC-8 for the 3 settings (accuracy).}
    \label{fig:rcc8-all-results}
\end{figure*}
\section{Experimental Setup}\label{sec:experiments}

\paragraph{Input Representation} In principle, the only contextual information needed to solve an instance of STaR is the composition table. \citet{khalid2025systematic} considered to what extent neuro-symbolic models were able to learn (and then systematically apply) this composition table from the training data provided. Here, we focus on a simpler setting, where we provide the composition table as part of the prompt. Our main focus is thus on whether LLMs and LRMs are able to follow the instructions and apply the composition rules in a systematic way. This allows us to evaluate models in a zero-shot fashion, or with a small number of in-context demonstrations (as well as evaluating fine-tuned models which should in principle be able to learn the composition table). We specify the composition table using a compact integer encoding (using powers of two; see the appendix for an example of the full prompt). The graph that defines a given problem instance is similarly encoded using integer labels.
The model is furthermore instructed to provide the answer using the same integer encoding. 
This is illustrated in Fig.~\ref{fig:illustration-figPrompt}.

\begin{figure*}[t!]
    \centering
    \includegraphics[width=0.95\linewidth]{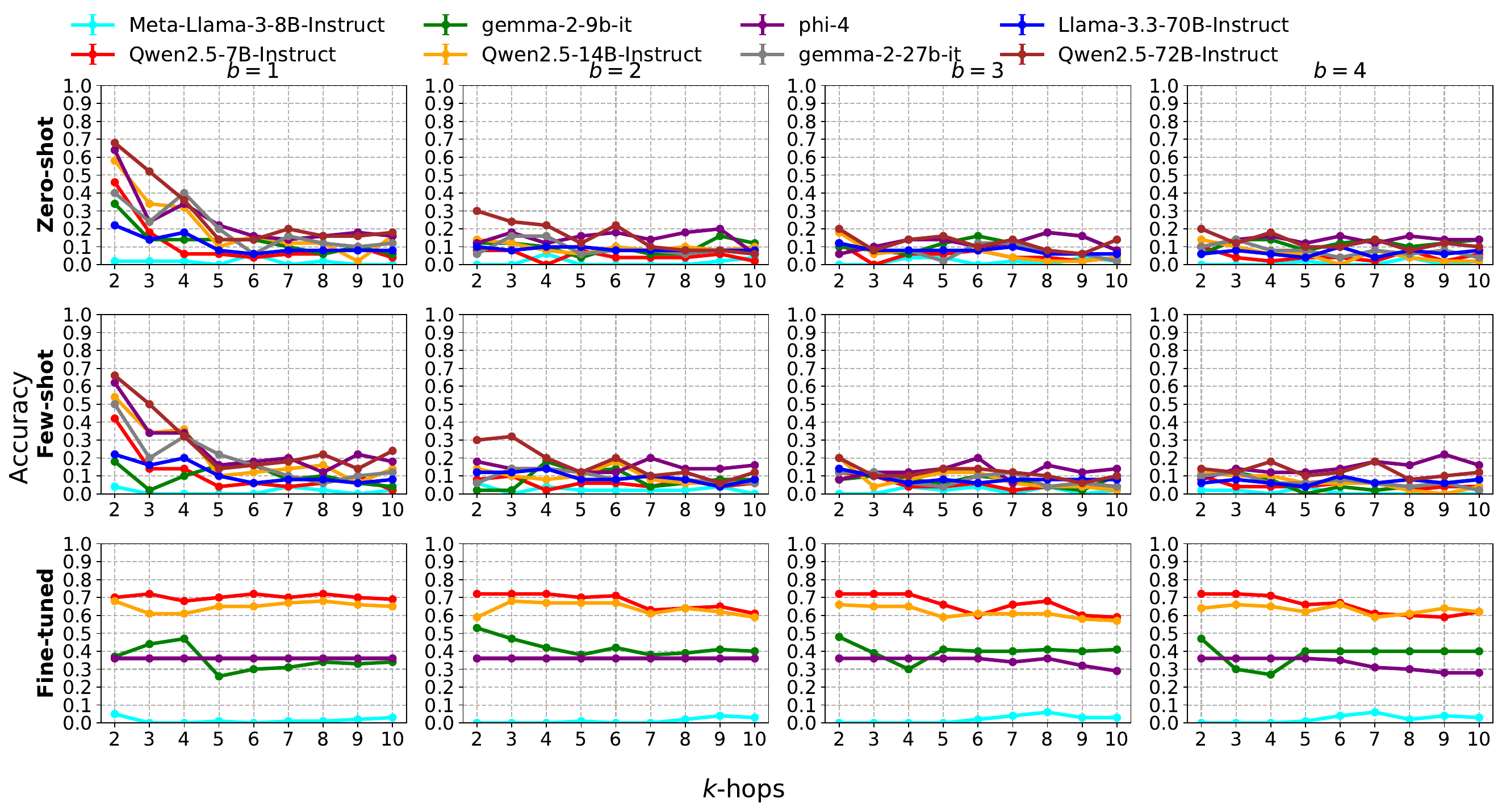}
    \caption{The results for the non-reasoning models on IA for the 3 settings (accuracy).}
    \label{fig:interval-all-results}
\end{figure*}

\paragraph{Evaluation Setup} To evaluate the models, for each combination of $(k,b)$, we use a uniform subsample of the full set of test problem instances for RCC-8 and IA. For RCC-8, each of the eight relations appears equally frequently as gold labels, meaning that the performance of naive baselines such as random guessing is at $1/8 = 0.125$. Similarly, the performance of naive baselines on IA is at $1/13\approx 0.076$.
We evaluate 2 types of models, LLMs (instruction tuned) and LRMs, on 3 distinct settings: (A) \textbf{Zero-shot}, (B) \textbf{Few-shot} and (C) \textbf{Fine-tuned}. Settings (A) and (B) evaluate the model's in-context learning and instruction following abilities. 
For the few-shot experiments, we provide 5 in-context demonstrations of the desired input/output pairs. For the experiments with fine-tuned models, we leverage the entire training set comprising 57600 and 93400 instances for RCC-8 and IA respectively. For testing, for settings (A) and (B) we use 500 test sample instances for RCC-8 and 100 for IA, for each combination of $k$ and $b$. We use 50 samples per $(k,b)$ configuration for setting (C) and the reasoning experiment. We use the following models: Llama-3 and Llama-3.3 \cite{grattafiori2024llama3herdmodels}, Qwen  \cite{qwen2025qwen25technicalreport}, Phi-4 \cite{abdin2024phi4technicalreport},  Gemma-2 \cite{gemmateam2024gemma2improvingopen} o3-mini \cite{openai2025competitiveprogramminglargereasoning}. The setup is summarized in Table~\ref{table:model-stats}. Further implementation details and data statistics are provided in App.~\ref{app:impl-details}.

\section{Results}\label{secResults}
Systematicity results are divided into two sections, first focussing on the non-reasoning models in Section \ref{secNonReasoningResults} (i.e.\ the standard LLMs), and then on the reasoning models in Section \ref{secResultsReasoningModels}.

\subsection{Non-reasoning Models}\label{secNonReasoningResults}
The results for RCC-8 are summarized in Figure \ref{fig:rcc8-all-results} and for IA in Figure \ref{fig:interval-all-results}. Broadly, both of these are similar. We therefore focus on RCC-8 below.

For the zero-shot experiments, all models perform close to random guessing for all but the simplest problem instances. Somewhat better results are observed only when $b\leq 2$ and $k\leq 4$. Qwen2.5-72B overall emerges as the strongest model. Its results remain clearly above random chance (although still very weak) for $b=1$ and $k\geq 5$. For lower values of $k$, gemma-2-9b and gemma-2-27b are the next best-peforming models. Interestingly, the much larger Llama-3.3-70B model performs poorly for low values of $k$, but performs the best for $b=2,k=10$, and similar to Qwen2.5-72B for $b=1, k=10$.

The results for the few-shot experiments are similar, with non-trivial performance only achieved for $k\leq 3$. Qwen2.5-72B performs consistently better than in the zero-shot case. 
The most interesting changes can be seen for $b=1$, where some of the smaller models now perform notably better, especially Qwen2.5-14B, gemma-2-9b and phi-4. Finally, the results for the fine-tuned models are much better. We can see a noticeable performance gap, with the Qwen models and gemma-2-27b  clearly outperforming the others. 
It is surprising to see that the performance for $b=2$, $b=3$ and $b=4$ is similar, despite the latter setting being much harder. We will come back to this point in Section \ref{secAnalysis}. In short, however, this is due to the fact that these models have learned to reliably predict some of the simplest relations, exploiting the \emph{trivial path} heurisitc. A path is trivial if there is a (s-t) path between the source (s) and tail (t) entities that only consists of $\mathsf{eq}$ and at most one other relation. Then the solution is either the identity or that non-identity relation.
The ability of these models to discover underlying principles, and reliably apply them in OOD settings is remarkable but it is clear that they are not capable of principled reasoning, as their performance on the hardest relations remains poor. The performance of all the considered models, even the best-performing fine-tuned models, remains far below that of state-of-the-art neuro-symbolic methods \cite{khalid2025systematic}, which achieve near-perfect results on these problem instances, despite having to learn the composition table from training examples.

\subsection{Reasoning Models}\label{secResultsReasoningModels}
\begin{table}[t]
    \centering
    \footnotesize
    \setlength\tabcolsep{6pt}
\begin{tabular}{llcccccc}
\toprule
 & Conf. & \multicolumn{2}{c}{o3-mini} & \multicolumn{2}{c}{Qwen 7B} & \multicolumn{2}{c}{Qwen 14B} \\
\cmidrule(lr){3-4} \cmidrule(lr){5-6} \cmidrule(lr){7-8}
      & $(k,b)$        & {Acc} & {F1}  & {Acc} & {F1}  & {Acc.} & {F1}  \\
\midrule
\parbox[t]{1mm}{\multirow{6}{*}{\rotatebox[origin=c]{90}{\textbf{RCC-8}}}}
&(9, 3) & 0.30 & 0.24 & 0.12 & 0.07 & 0.06 & 0.05 \\
&(9, 2) & 0.48 & 0.38 & 0.06 & 0.02 & 0.26 & 0.23 \\
&(9, 1) & 0.90 & 0.85 & 0.08 & 0.07 & 0.20 & 0.15 \\
&(8, 4) & 0.44 & 0.35 & 0.10 & 0.08 & 0.16 & 0.12 \\
&(8, 3) & 0.56 & 0.52 & 0.12 & 0.11 & 0.14 & 0.10 \\
&(5, 2) & 0.68 & 0.63 & 0.12 & 0.07 & 0.24 & 0.19 \\
\midrule
\parbox[t]{1mm}{\multirow{6}{*}{\rotatebox[origin=c]{90}{\textbf{IA}}}}
&(9, 3) & 0.30 & 0.29 & 0.04 & 0.03 & 0.10 & 0.10 \\
&(9, 2) & 0.44 & 0.42 & 0.06 & 0.04 & 0.22 & 0.18 \\
&(9, 1) & 0.78 & 0.74 & 0.20 & 0.15 & 0.14 & 0.09 \\
&(8, 4) & 0.36 & 0.30 & 0.04 & 0.06 & 0.12 & 0.07 \\
&(8, 3) & 0.34 & 0.36 & 0.04 & 0.03 & 0.14 & 0.07 \\
&(5, 2) & 0.56 & 0.52 & 0.04 & 0.03 & 0.18 & 0.11 \\
\bottomrule
\end{tabular}
    \caption{Zero-shot (setting (A)) results for the reasoning models on the STaR benchmark. The Qwen models are distilled R1 models which were run locally. The accuracies and macro F1 scores are reported for a sample of test configurations due to API resource constraints. }
    \label{table:reasoning-results}
\end{table}

\begin{table}[t]
\footnotesize
    \centering
\begin{tabular}{lrcccc}
\toprule
   & Label & Pr.  & Re.  & F1.  & Count \\
\midrule
\parbox[t]{1mm}{\multirow{8}{*}{\rotatebox[origin=c]{90}{\textbf{RCC-8}}}}
&\texttt{DC}    & 0.14          & 0.31          & 0.20          & 13 \\
&\texttt{EC}    & 0.43          & 0.25          & 0.32          & 12 \\
&\texttt{PD}    & 0.14          & 0.18          & 0.16          & 11 \\
&\texttt{TPP}   & \textbf{1.00} & 0.09          & 0.17          & 11 \\
&\texttt{NTPP}  & 0.00          & 0.00          & 0.00          & \textbf{17} \\
&\texttt{TPPI}  & 0.72          & \textbf{1.00} & 0.84          & 13 \\
&\texttt{NTPPI} & 0.68          & \textbf{1.00} & 0.81          & 13 \\
&\texttt{EQ}    & \textbf{1.00} & \textbf{1.00} & \textbf{1.00} & 10 \\
\midrule
\parbox[t]{1mm}{\multirow{13}{*}{\rotatebox[origin=c]{90}{\textbf{IA}}}}
&\texttt{=}     & 0.14          & 0.83          & 0.24          & 6  \\
&\texttt{<}     & 0.00          & 0.00          & 0.00          & 4  \\
&\texttt{>}     & 0.00          & 0.00          & 0.00          & 9  \\
&\texttt{d}     & \textbf{1.00} & 0.10          & 0.18          & 10 \\
&\texttt{di}    & 0.00          & 0.00          & 0.00          & 9  \\
&\texttt{o}     & \textbf{1.00} & 0.57          & 0.73          & 7  \\
&\texttt{oi}    & \textbf{1.00} & \textbf{1.00} & \textbf{1.00} & 5  \\
&\texttt{m}     & \textbf{1.00} & \textbf{1.00} & \textbf{1.00} & 9  \\
&\texttt{mi}    & \textbf{1.00} & 0.67          & 0.80          & 6  \\
&\texttt{s}     & \textbf{1.00} & \textbf{1.00} & \textbf{1.00} & 9  \\
&\texttt{si}    & \textbf{1.00} & \textbf{1.00} & \textbf{1.00} & 8  \\
&\texttt{f}     & \textbf{1.00} & 0.83          & 0.91          & 6  \\
&\texttt{fi}    & \textbf{1.00} & \textbf{1.00} & \textbf{1.00} & 12 \\
\bottomrule
\end{tabular}
\caption{Fine-grained breakdown of classification scores for the $k=9, b=2$ dataset configuration for the fine-tuned Qwen2.5-14B LLM.  We sample 50 points randomly from each STaR dataset. }
    \label{table:reasoning-results-finetunedQwen}
\end{table}

\begin{table}[t]
\footnotesize
    \centering
\begin{tabular}{lrcccc}
\toprule
   & Label & Pr.  & Re.  & F1.  & Count \\
\midrule
\parbox[t]{1mm}{\multirow{8}{*}{\rotatebox[origin=c]{90}{\textbf{RCC-8}}}}
&\texttt{DC}                 & 0.69          & 0.90          & \textbf{0.78} & 10 \\
&\texttt{EC}                 & 0.50          & \textbf{1.00} & 0.67          & 3  \\
&\texttt{PD}                 & 0.43          & 0.27          & 0.33          & 11 \\
&\texttt{TPP}                & 0.33          & 0.44          & 0.38          & 9  \\
&\texttt{NTPP}               & \textbf{1.00} & 0.20          & 0.33          & 5  \\
&\texttt{TPPI}               & 0.00          & 0.00          & 0.00          & 2  \\
&\texttt{NTPPI}              & 0.50          & 0.25          & 0.33          & 4  \\
&\texttt{EQ}                 & 0.75          & 0.50          & 0.60          & 6  \\
\midrule
\parbox[t]{1mm}{\multirow{13}{*}{\rotatebox[origin=c]{90}{\textbf{IA}}}}
&\texttt{=}                  & 0.50          & 0.17          & 0.25          & 6  \\
&\texttt{<}                  & 0.10          & \textbf{1.00} & 0.18          & 1  \\
&\texttt{>}                  & 0.83          & \textbf{1.00} & \textbf{0.91} & 5  \\
&\texttt{d}                  & 0.50          & 0.60          & 0.55          & 5  \\
&\texttt{di}                 & 0.67          & 0.50          & 0.57          & 4  \\
&\texttt{o}                  & 0.00          & 0.00          & 0.00          & 2  \\
&\texttt{oi}                 & 0.75          & \textbf{1.00} & 0.86          & 3  \\
&\texttt{m}                  & \textbf{1.00} & 0.50          & 0.67          & 4  \\
&\texttt{mi}                 & \textbf{1.00} & 0.33          & 0.50          & 3  \\
&\texttt{s}                  & \textbf{1.00} & 0.20          & 0.33          & 5  \\
&\texttt{si}                 & \textbf{1.00} & 0.25          & 0.40          & 4  \\
&\texttt{f}                  & 0.33          & 0.50          & 0.40          & 2  \\
&\texttt{fi}                 & \textbf{1.00} & 0.17          & 0.29          & 6  \\
\bottomrule
\end{tabular}
\caption{Fine-grained breakdown of classification scores for the $k=9, b=2$ dataset configuration for the o3-mini LRM.  We sample 50 points randomly from each STaR dataset.  }
    \label{table:reasoning-results-o3mini}    
\end{table}

For the reasoning models, we focus on the zero-shot evaluation setting. The results are summarized in Table \ref{table:reasoning-results}. Note that we only include results for a sample of all $(k,b)$ configurations due to the much higher cost that is involved in using these models. 

Compared to the non-reasoning models without fine-tuning, the performance of o3-mini \cite{openai2025competitiveprogramminglargereasoning} is remarkably strong. The setting with $b=1$ is intuitively well-aligned with the chain-of-thought process. Accordingly, we can see that the model performs well for $b=1$, even with $k=9$, achieving an accuracy of 0.9, which is substantially higher than what any of the fine-tuned models has achieved. However, for $b\geq 2$ the results quickly deteriorate. Interestingly, this behavior is qualitatively different from that of the fine-tuned models. Where the fine-tuned models have learned to identify trivial path relations, o3-mini seems capable of interpreting the composition table and systematically applying it to a single reasoning path (although not with perfect accuracy, even for $b=1$). For $b\geq 2$, the disjunctive nature of the reasoning problem proves problematic, suggesting that the model is limited in its capacity to generalize to unseen reasoning tasks. For the distilled Deepseek-R1 models \cite{guo2025deepseek}, the results are below random chance for all settings where $b\geq 2$. For $k=9$ and $b=1$, the results are above random chance (except for Qwen 7B on RCC-8), but not meaningfully better than the non-reasoning models in the zero-shot setting.

\section{Analysis}\label{secAnalysis}

\subsection{Fine-grained Classification Breakdown}

In Section \ref{secResults}, we already saw that the behavior of the fine-tuned LLMs, on the one hand, and o3-mini, on the other hand, was qualitatively different. To further analyze this, Table \ref{table:reasoning-results-finetunedQwen} shows a breakdown of the results per relation type, for one of the best-performing fine-tuned models (Qwen2.5-14B). Table \ref{table:reasoning-results-o3mini} shows the same breakdown for o3-mini. In both tables, we focus on the case where $k=9$ and $b=2$. Focusing on Table \ref{table:reasoning-results-finetunedQwen} first, for RCC-8 we can see that the fine-tuned Qwen2.5-14B model achieves perfect results on $\mathsf{eq}$, which can be explained by the fact that this relation can only be predicted if there is a trivial (s-t) path. For $\mathsf{ntppi}$ and $\mathsf{tppi}$, the model was able to exploit a similar insight. The performance on the other relations, however, is much worse, although still better than random chance (except for $\mathsf{ntpp}$). For IA, we can see a similar pattern. Some of the relations are easier to predict, with the model achieving perfect results on several relations: $\mathsf{oi}$, $\mathsf{m}$, $\mathsf{s}$, $\mathsf{si}$ and $\mathsf{fi}$. However, for other relations, the results are very poor. This again shows that the model was able to learn some ``tricks'' that allow it to reliably predict some of the easier relations, even on out-of-distribution settings, while at the same time failing to apply the rules from the composition table in a systematic way.

The results for o3-mini in Table \ref{table:reasoning-results-o3mini} paint a dramatically different picture. First, note that o3-mini does not achieve perfect results on any of the relations. This shows that it was not able to leverage domain-specific insights (such as the idea that $\mathsf{eq}$ can only be predicted if there is a chain of $\mathsf{eq}$-relations). On the other hand, the model achieves non-trivial results for almost all the relations. This suggests that the correct predictions are due to the ability of the model to follow the instructions from the composition table in a somewhat systematic, albeit error-prone way.

\begin{figure}
    \centering
    \includegraphics[width=1\linewidth]{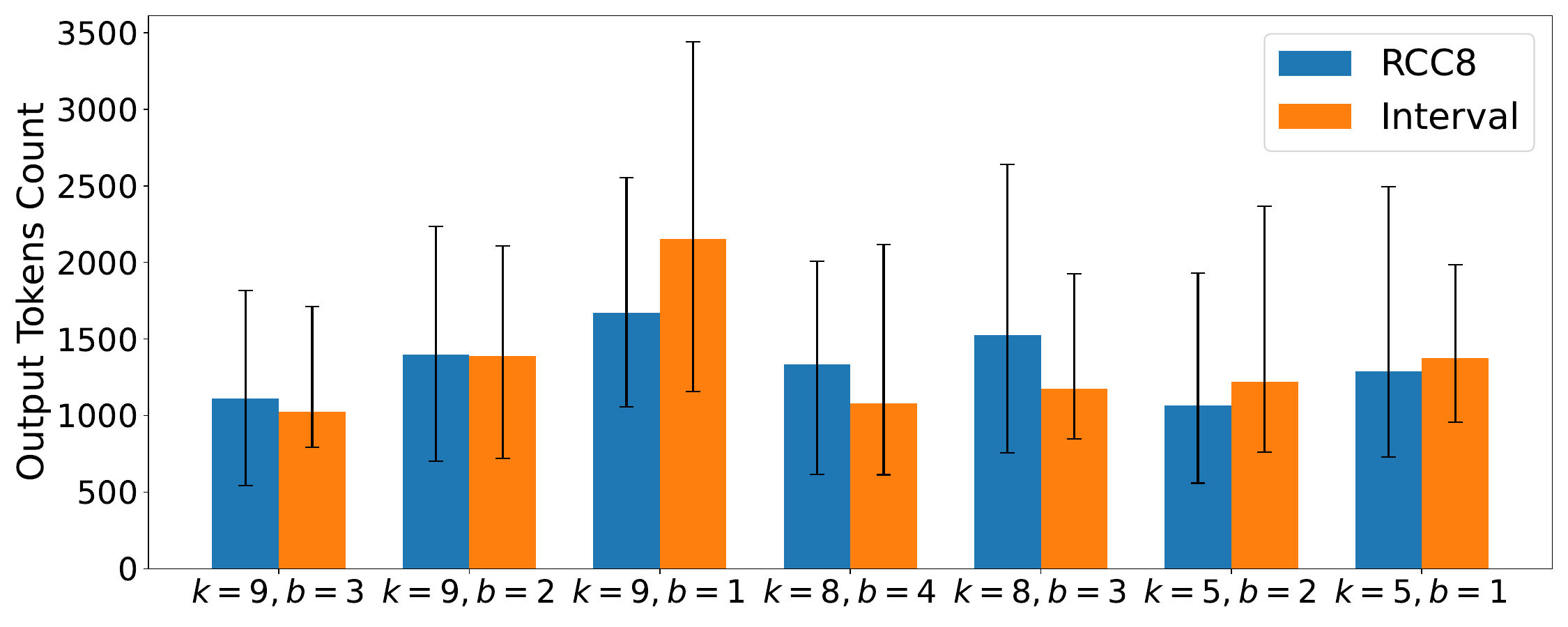}
    \caption{The median number of output tokens with the interquartile range for the Qwen 7B reasoning model for the same dataset splits as in Table~\ref{table:reasoning-results}. The number of maximum tokens was set to 8192. }
    \label{fig:cot-output-tokens}    
 \end{figure}
 
\subsection{CoT Analysis}
Reasoning models can adapt the number of output tokens, i.e.\ the amount of test-time compute, based on the difficulty of a given problem instance. To analyze this aspect, Figure \ref{fig:cot-output-tokens} shows the number of output tokens that were generated by the Qwen 7B reasoning model.  Note that we cannot do this analysis for o3-mini as the intermediate reasoning process is hidden for this model. Counterintuitively, the analysis in Figure \ref{fig:cot-output-tokens} reveals that the number of output tokens goes down, as the number of paths $b$ increases, for all the considered values of $k$. This seems to suggest that the model is aware of its limitations on these problem instances, giving up the reasoning process more quickly. In contrast, we can see that considerably more output tokens were used for $k=9, b=1$ than for $k=5, b=1$, which further supports our hypothesis that single-path problem instances are more natural for chain-of-thought based reasoning.
\begin{figure}[t]
    \centering
    \includegraphics[width=0.75\linewidth]{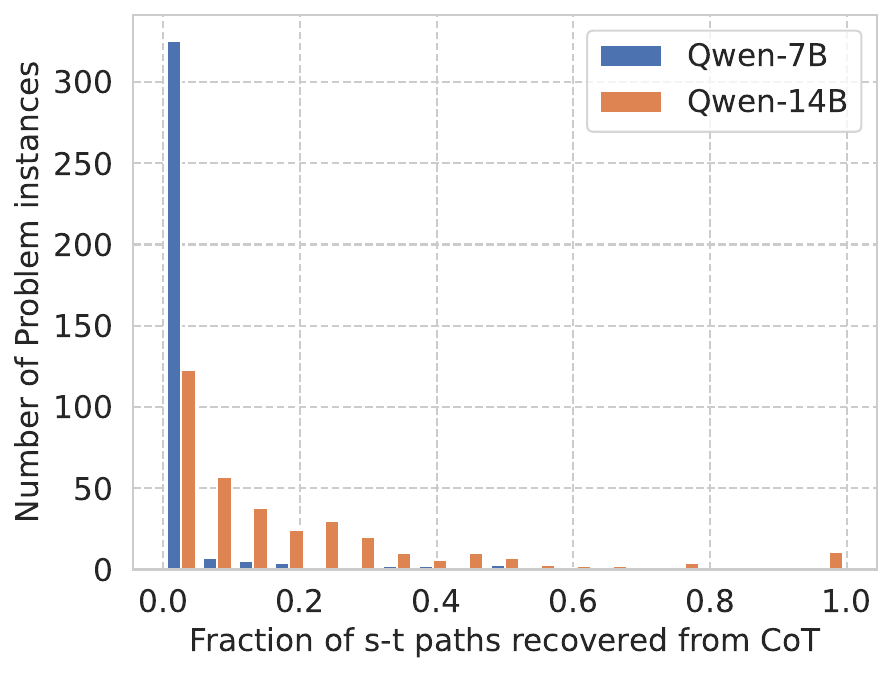}
    \caption{Fraction of source-to-tail paths recovered from the model's CoT for IA.}
    \label{fig:path-recovery-on-ia}
\end{figure}
\begin{figure*}[t]
    \centering
    \includegraphics[width=1.\linewidth]{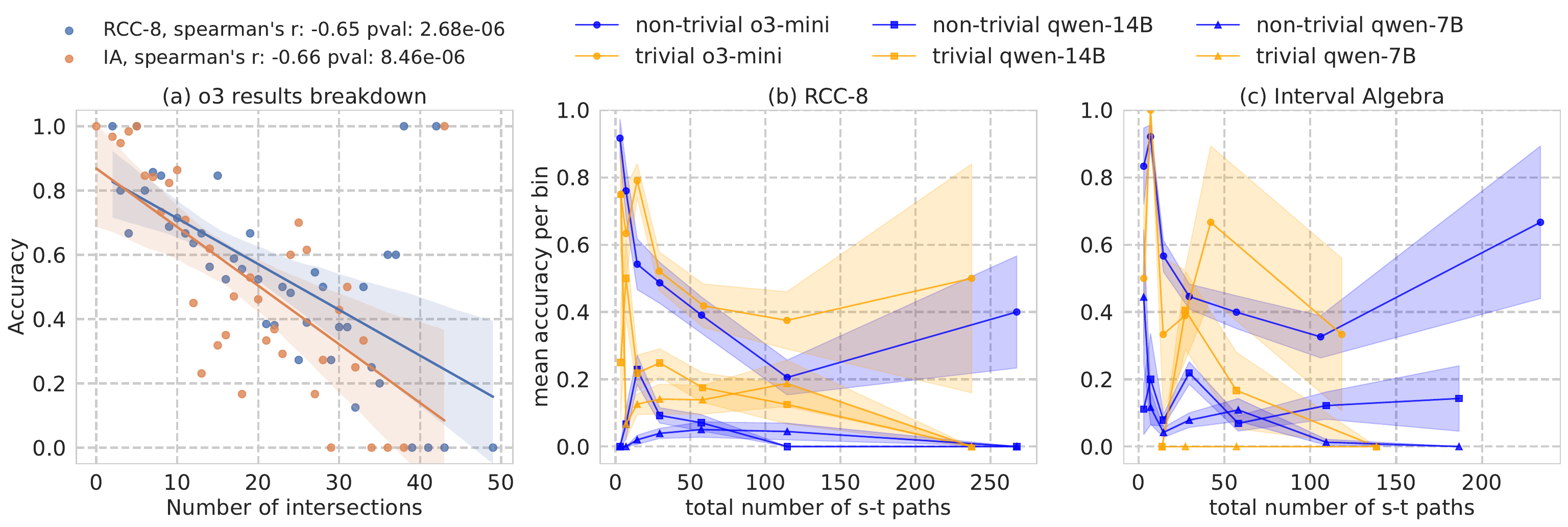}
    \caption{LRMs are shallow Algebraic Closure Algorithm (ACA) simulators. (a) o3-mini's performance on both RCC-8 and IA datasets degrades approximately linearly as a function of the number of intersection operations in a problem instance, which are required whenever the in-degree of a node in the graph is greater than 1. (b)-(c) o3-mini, R1 distilled Qwen-7B and Qwen-14B noisily degrade in performance as the number of source-tail paths in the problem instance increases. Performance scaling with model size is also observed. Remarkably, the models, increasingly with size, zero-shot exploit the trivial path heuristic for solving STaR problems. Error bars are $\pm1\sigma$.}
    \label{fig:shallow-algo-simulator-1}
\end{figure*}
\subsection{Shallow Algebraic Closure Algorithm Simulation}
To solve disjunctive reasoning, the model needs to simulate a disjunctive reasoning algorithm, the algebraic closure algorithm (ACA) \cite{algebraic-closure} where multiple possible solutions are refined iteratively during graph traversal. ACA is novel compared to reasoning algorithms for linearizable computation graphs \cite{dziri2023faith} or constraint satisfiability problems \cite{lin2025zebralogic}. Importantly, the intermediate nodes store partial solutions that are atomic whereas for disjunctive reasoning the nodes need to contain multiple possible solutions or sets.

ACA consists of 3 basic operations: relational composition, union and intersection, where the last two are operations on sets of possible relations. In addition, graph traversal is necessary to find paths between two nodes e.g.\ by simulating the Bellman-Ford algorithm. Our findings show that reasoning models are shallow ACA simulators by looking at the trend in performance as a function of these basic operations. All LRMs attempt to solve STaR-type problems using path-based reasoning which necessitates enumerating all possible paths between the source and tail nodes in the query edge. 

Firstly, we analyze the CoT of open LRMs to quantify the fraction of unique source-to-tail (s-t) paths that is recovered by the models. Figure~\ref{fig:path-recovery-on-ia} shows, for Qwen 7B and 14B models on IA, an exponential decline in the coverage of paths per problem instance, in spite of being alloted ample CoT tokens (cf.\ App~\ref{app:cot-example} for an example CoT and the companion RCC-8 analysis). Manually inspecting some CoT summaries from o3-mini (where full CoT is inaccessible) confirms similar behavior. This implies that the reasoning algorithms cannot properly simulate graph traversal, in line with search-related findings~\cite{transformers-struggle-search}. 

Secondly, we quantify the number of intersections required per STaR problem instance by adding all in-degrees of a node in the graph minus 1. This is a strong predictor of problem difficulty as it directly quantifies its disjunctive multi-path nature. A clear linear trend in the decline in performance of o3-mini for RCC-8 and IA as a function of the number of intersections is shown in Figure~\ref{fig:shallow-algo-simulator-1}(a). Union and composition operations are correlated and occur with each edge-wise composition in the graph so it is harder to quantify their quality but similar noisier trends are observed in App.~\ref{app:aca-analysis}. 

Thirdly, we confirm explicitly that all models zero-shot exploit the trivial path heuristic to solve problem instances that avoid a full computation. From the CoTs, since all the models reason by considering full s-t paths but are unable to enumerate them all, they are unable to always exploit this heuristic, making use of it only when it is seen. It is remarkable that LRMs are able to exploit this heuristics, as they were not trained on RCC-8 or IA problem instances (to the best of our knowledge). There is a separation in performance with respect to the presence of trivial paths in problem instances for all reasoning models on RCC-8 and IA and is shown in Figure~\ref{fig:shallow-algo-simulator-1}(b)-(c). Moreover, performance improvement with respect to size scaling is also clearly observable.

\section{Conclusions}
We have studied the performance of recent LLMs and so-called Large Reasoning Models (LRMs) on a challenging benchmark involving qualitative spatial and temporal reasoning problems. This analysis allows us test the abilities of models on a different style of reasoning than those that are typically considered, and crucially, than those that are used for training LRMs. The setting requires composing relations, using rules that are specified in a composition table. A particular challenge arises because multiple ``reasoning paths'' need to be combined to arrive at the final answer, which is harder to capture using a chain-of-thought process. 

Several insights arise from our analysis. While LLMs perform poorly in zero-shot and few-shot settings, fine-tuned LLMs achieved notably better results. However, further analysis shows that this is because fine-tuned models achieve near-perfect results on some of the easier test instances, i.e.\ relations that can be predicted by relying on simple rules and heuristics, rather than a systematic application of the composition table. In particular, these models still perform poorly on problem instances that require multi-path reasoning. As far as LRMs are concerned, o3-mini performs much better than LLMs in zero-shot and few-shot settings, but does not overall improve on the performance of fine-tuned LLMs. Interestingly, the behavior of the fine-tuned LLMs and o3-mini is qualitatively different. Indeed, o3-mini seems to rely more on an error-prone, but systematic application of the rules from the composition table, achieving strong results for problems involving only a single reasoning path. However, when multiple reasoning paths need to be combined, its performance deteriorates quickly. We further conduct a behavioral analysis of how the LRMs perform and find that they are shallow disjunctive reasoning algorithm simulators due to their inability to properly simulate crucial steps like graph traversal and intersection.

These results suggest that LRMs, despite demonstrating improved reasoning, are still limited in terms of their ability to generalize to previously unseen reasoning tasks.

\section*{Acknowledgments}
This work was supported by the EPSRC grant EP/W003309/1.

\section*{Limitations}
The state-of-the-art in reasoning models is still quickly changing, and any conclusions that can be drawn from current models, such as o3-mini, may quickly become obsolete as newer models are released. A key question, which remains unanswered, is whether reasoning models can be designed that generalize to previously unseen reasoning tasks. Furthermore, while we have advocated the use of temporal and spatial reasoning, further analysis is needed to test the reasoning abilities of current models on a broader range of problems, and to better understand their failure modes more generally. In terms of the considered models, we have focused our analysis on open-source models that can be run locally (with the exception of o3-mini), and quantization was used to make this possible. It is possible that fine-tuning larger models may lead to better results. 
%
Finally, 
only a limited set of $(k,b)$ configurations was used to evaluate the reasoning models due to compute constraints. 


\bibliography{custom}

\appendix

\begin{figure}[t!]
    \centering 
    \includegraphics[width=220pt]{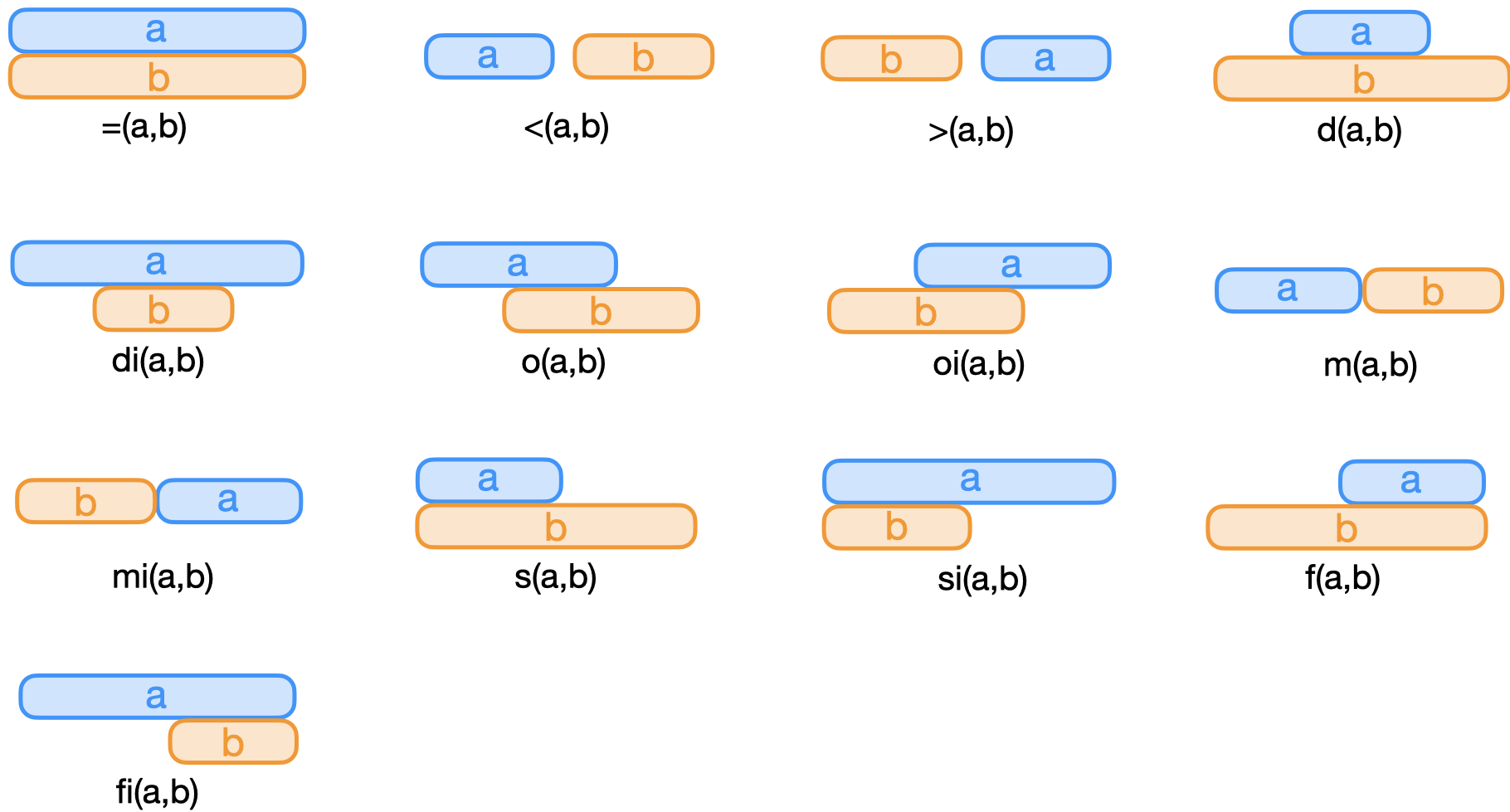}
\caption{Illustration of the IA relations. }
\label{fig:illustration-figIA}
\end{figure}

\begin{table*}
\centering
\scriptsize
\begin{tabular}{|m{15pt}||m{39pt}|m{39pt}|m{39pt}|m{39pt}|m{44pt}|m{39pt}|m{39pt}|}
\hline
				    & $\dc$ & $\ec$ & $\po$ & $\tpp$ & $\ntpp$ & $\tppi$ & $\ntppi$ \\
\hline
\hline
$\dc$       & $\mathcal{R}_8$   & $\dc$, $\ec$, $\po$, $\tpp$, $\ntpp$   & $\dc$, $\ec$, $\po$, $\tpp$, $\ntpp$        & $\dc$, $\ec$, $\po$, $\tpp$, $\ntpp$  & $\dc$, $\ec$, $\po$, $\tpp$, $\ntpp$  &  $\dc$    & $\dc$ \\
\hline
$\ec$           & $\dc$, $\ec$, $\po$, $\tppi$, $\ntppi$       & $\dc$, $\ec$, $\po$, $\tpp$, $\tppi$, $\eq$      & $\dc$, $\ec$, $\po$, $\tpp$, $\ntpp$ & $\ec$, $\po$, $\tpp$, $\ntpp$ & $\po$, $\tpp$, $\ntpp$ & $\dc$, $\ec$ & $\dc$ \\             
\hline
$\po$        & $\dc$, $\ec$, $\po$, $\tppi$, $\ntppi$   & $\dc$, $\ec$, $\po$, $\tppi$, $\ntppi$        & $\mathcal{R}_8$   & $\po$, $\tpp$, $\ntpp$  & $\po$, $\tpp$, $\ntpp$ & $\dc$, $\ec$, $\po$, $\tppi$, $\ntppi$       & $\dc$, $\ec$, $\po$, $\tppi$, $\ntppi$   \\           
\hline
$\tpp$       & $\dc$ & $\dc$, $\ec$    & $\dc$, $\ec$, $\po$, $\tpp$, $\ntpp$  & $\tpp$, $\ntpp$   & $\ntpp$ & $\dc$, $\ec$, $\po$, $\tpp$, $\tppi$, $\eq$      & $\dc$, $\ec$, $\po$, $\tppi$, $\ntppi$  \\           
\hline
$\ntpp$      & $\dc$ & $\dc$    & $\dc$, $\ec$, $\po$, $\tpp$, $\ntpp$  & $\ntpp$ & $\ntpp$ & $\dc$, $\ec$, $\po$, $\tpp$, $\ntpp$  &  $\mathcal{R}_8$ \\
\hline
$\tppi$     & $\dc$, $\ec$, $\po$, $\tppi$, $\ntppi$        & $\ec$, $\po$, $\tppi$, $\ntppi$    & $\po$, $\tppi$, $\ntppi$    & $\po$, $\eq$, $\tpp$, $\tppi$     & $\po$, $\tpp$, $\ntpp$ & $\tppi$, $\ntppi$ & $\ntppi$ \\            
\hline
$\ntppi$  & $\dc$, $\ec$, $\po$, $\tppi$, $\ntppi$       & $\po$, $\tppi$, $\ntppi$ & $\po$, $\tppi$, $\ntppi$        & $\po$, $\tppi$, $\ntppi$        & $\po$, $\tppi$, $\tpp$, $\ntpp$, $\ntppi$, $\eq$ & $\ntppi$ & $\ntppi$\\      
\hline
\end{tabular}
\caption{RCC-8 composition table \cite{rcc8}, excluding the trivial composition with $\eq$. We write $\mathcal{R}_8$ for the trivial case, where the composition consists of all eight relations.  \label{tableRCC8composition}}
\end{table*}

\begin{table*}[t]
\centering
\scriptsize
\centerline{\begin{tabular}{|m{15pt}||m{18pt}|m{18pt}|m{18pt}|m{18pt}|m{18pt}|m{18pt}|m{18pt}|m{18pt}|m{18pt}|m{18pt}|m{18pt}|m{18pt}|}
\hline
				    & $<$ & $>$ & $\intervald$ & $\intervaldi$ & $\intervalo$ & $\intervaloi$ & $\intervalm$ & $\intervalmi$ & $\intervals$ & $\intervalsi$ & $\intervalf$ & $\intervalfi$   \\
\hline
\hline
$<$ 
& $<$  &  & $<,\intervalo$, $\intervalm$, $\intervald$, $\intervals$ & $<$ & $<$ & $<,\intervalo$, $\intervalm$, $\intervald$, $\intervals$ & $<$ & $<,\intervalo$, $\intervalm$, $\intervald$, $\intervals$ & $<$ & $<$ & $<,\intervalo$, $\intervalm$, $\intervald$, $\intervals$ & $<$ \\
\hline
$>$   
& & $>$ & $>$, $\intervaloi$, $\intervalmi$, $\intervald$, $\intervalf$ & $>$ 
& $>$, $\intervaloi$, $\intervalmi$, $\intervald$, $\intervalf$ & $>$ & $>$, $\intervaloi$, $\intervalmi$, $\intervald$, $\intervalf$ & $>$ & $>$, $\intervaloi$, $\intervalmi$, $\intervald$, $\intervalf$ & $>$ & $>$ & $>$  \\             
\hline
$\intervald$
& $<$ & $>$ & $\intervald$ & & $<$, $\intervalo$, $\intervalm$, $\intervald$, $\intervals$ & $>$, $\intervaloi$, $\intervalmi$, $\intervald$, $\intervalf$ & $<$ & $>$ & $\intervald$ & $>$, $\intervaloi$, $\intervalmi$, $\intervald$, $\intervalf$ & $\intervald$ &  $<$, $\intervalo$, $\intervalm$, $\intervald$, $\intervals$   
\\           
\hline
$\intervaldi$       
& $<$, $\intervalo$, $\intervalm$, $\intervaldi$, $\intervalfi$ & 
$>$, $\intervaloi$, $\intervaldi$, $\intervalmi$, $\intervalsi$ & $\intervalo$, $\intervaloi$, $\intervald$, $\intervals$, $\intervalf$, $\intervaldi$, $\intervalsi$, $\intervalfi$, $=$ & $\intervaldi$ & $\intervalo$, $\intervaldi$, $\intervalfi$ & $\intervaloi$, $\intervaldi$, $\intervalsi$ & $\intervalo$, $\intervaldi$, $\intervalfi$ & $\intervaloi$, $\intervaldi$, $\intervalsi$ & $\intervalo$, $\intervaldi$, $\intervalfi$ & $\intervaldi$ & $\intervaloi$, $\intervaldi$, $\intervalsi$ & $\intervaldi$ 
\\           
\hline
$\intervalo$      
& $<$ & $>$, $\intervaloi$, $\intervaldi$, $\intervalmi$, $\intervalsi$ 
& $\intervalo$, $\intervald$, $\intervals$ & $<$, $\intervalo$, $\intervalm$, $\intervaldi$, $\intervalfi$ & $<$, $\intervalo$, $\intervalm$ & $\intervalo$, $\intervaloi$, $\intervald$, $\intervals$, $\intervalf$, $\intervaldi$, $\intervalsi$, $\intervalfi$, $=$ & $<$ & $\intervaloi$, $\intervaldi$, $\intervalsi$ & $\intervalo$ & $\intervalo$, $\intervaldi$, $\intervalfi$ & $\intervalo$, $\intervald$, $\intervals$ & $<$, $\intervalo$, $\intervalm$   
\\
\hline
$\intervaloi$     
&
$<$, $\intervalo$, $\intervalm$, $\intervaldi$, $\intervalfi$ & $>$ & $\intervaloi$, $\intervald$, $\intervalf$ & $>$, $\intervaloi$, $\intervalmi$, $\intervaldi$, $\intervalsi$ & $\intervalo$, $\intervaloi$, $\intervald$, $\intervaldi$, $\intervals$, $\intervalsi$, $\intervalf$, $\intervalfi$, $=$ & $>$, $\intervaloi$, $\intervalmi$ & $\intervalo$, $\intervaldi$, $\intervalfi$ & $>$ & $\intervaloi$, $\intervald$, $\intervalf$ & $\intervaloi$, $>$, $\intervalmi$ & $\intervaloi$ & $\intervaloi$, $\intervaldi$, $\intervalsi$
\\            
\hline
$\intervalm$  & 
$<$ & $>$, $\intervaloi$, $\intervaldi$, $\intervalmi$, $\intervalsi$ & $\intervalo$, $\intervald$, $\intervals$ & $<$ & $<$ & $\intervalo$, $\intervald$, $\intervals$ & $<$ & $\intervalf$, $\intervalfi$, $=$ & $\intervalm$ & $\intervalm$ & $\intervald$, $\intervals$, $\intervalo$ & $<$
\\      
\hline
$\intervalmi$  & 
$<$, $\intervalo$, $\intervalm$, $\intervaldi$, $\intervalfi$ & $>$ & $\intervaloi$, $\intervald$, $\intervalf$ & $>$ & $\intervaloi$, $\intervald$, $\intervalf$ & $>$ & $\intervals$, $\intervalsi$, $=$ & $>$ & $\intervald$, $\intervalf$, $\intervaloi$ & $>$ & $\intervalmi$ & $\intervalmi$
\\      
\hline
$\intervals$  
& 
$<$ & $>$ & $\intervald$ & $<$, $\intervalo$, $\intervalm$, $\intervaldi$, $\intervalfi$ & $<$, $\intervalo$, $\intervalm$ & $\intervaloi$, $\intervald$, $\intervalf$ & $<$ & $\intervalmi$ & $\intervals$ & $\intervals$, $\intervalsi$, $=$ & $\intervald$ & $<$, $\intervalm$, $\intervalo$
\\      
\hline
$\intervalsi$  
& 
$<$, $\intervalo$, $\intervalm$, $\intervaldi$, $\intervalfi$ & $>$ & $\intervaloi$, $\intervald$, $\intervalf$ & $\intervaldi$ & $\intervalo$, $\intervaldi$, $\intervalfi$ & $\intervaloi$ & $\intervalo$, $\intervaldi$, $\intervalfi$ & $\intervalmi$ & $\intervals$, $\intervalsi$, $=$ & $\intervalsi$ & $\intervaloi$ & $\intervaldi$ 
\\      
\hline
$\intervalf$  
& 
$<$ & $>$ & $\intervald$ & $>$, $\intervaloi$, $\intervalmi$, $\intervaldi$, $\intervalsi$ & $\intervalo$, $\intervald$, $\intervals$ & $>$, $\intervaloi$, $\intervalmi$ & $\intervalm$ & $>$ & $\intervald$ & $>$, $\intervaloi$, $\intervalmi$ & $\intervalf$ & $\intervalf$, $\intervalfi$, $=$
\\      
\hline
$\intervalfi$  
& 
$<$ & $>$, $\intervaloi$, $\intervaldi$, $\intervalmi$, $\intervalsi$ & $\intervalo$, $\intervald$, $\intervals$ & $\intervaldi$ & $\intervalo$ & $\intervaloi$, $\intervaldi$, $\intervalsi$ & $\intervalm$ & $\intervalsi$, $\intervaloi$, $\intervaldi$ & $\intervalo$ & $\intervaldi$ & $\intervalf$, $\intervalfi$, $=$ & $\intervalfi$
\\      
\hline
\end{tabular}}
\caption{Allen's interval algebra composition table \citep{interval}, excluding the trivial composition with $=$.  \label{table-interval-composition}}
\end{table*}

\section{Details on RCC-8 and IA}
Figure \ref{fig:illustration-figIA} provides an illustration of the 13 relations of the interval algebra. The composition tables for RCC-8 and IA are shown respectively in Tables \ref{tableRCC8composition} and \ref{table-interval-composition}. To illustrate how reasoning with these calculi works, suppose we are given the following facts:
\begin{align*}
&\ec(a,b)
&&\ntpp(b,c)
&&\po(a,d)
&&\ec(d,c)
\end{align*}
Using the composition table, from $\ec(a,b)$ and $\ntpp(b,c)$, we know that the following must hold:
$$
\po(a,c) \vee \tpp(a,c) \vee \ntpp(a,c)
$$
Similarly, from $\po(a,d)$ and $\ec(d,c)$, we know that the following must hold:
$$
\dc(a{,}c) \vee \ec(a{,}c) \vee \po(a{,}c) \vee \tppi(a{,}c) \vee \ntppi(a{,}c)
$$
Since it is not possible for more than one relation to hold between $a$ and $c$, the only possibility is that $\po(a,c)$ holds. 

In general, sound and complete reasoning in RCC-8 and IA is possible by using the algebraic closure algorithm (for the case where the initial information does not contain any disjunctions). This algorithm amounts to maintaining, for each pair of entities, a set of possible relations. These sets are iteratively refined by applying composition rules, until convergence. The algorithm runs in cubic time. 
The problem instances in the StaR benchmark are simpler than general RCC-8 and IA problems. For these instances, it always suffices to consider the paths between the designated entities $h$ and $t$. Each path gives rise to a set of candidate relations, and the final answer is obtained by taking the intersection of these sets. The complexity of reasoning is thus linear in the number of entities. This ensures that the considered models should, in principle, be powerful enough to solve the problem instances, even for larger problems, and without needing an excessive number of output tokens for the LRMs.

\section{Implementation Details}\label{app:impl-details}
\subsection{Compute resources}
All relevant hyperparameters were tuned using grid search, as detailed below. All experiments were conducted using RTX 4090 and RTX 6000 Ada NVIDIA GPUs. For the small models, the results for all $(k,b)$ configurations, for the zero-shot, few-shot and fine-tuned settings, can be obtained in around 6-8 hours per model. For the large 70B models at 4-bit quantization, with a smaller sample size of 50 instances per $(k,b)$ configuration, a single full run (i.e.\ 24 $(k,b)$ configurations) takes around 1 day. We use the unsloth library \cite{unsloth} for fine-tuning all models with 4-bit quantization and the transformers library for downloading the weights and running all the open-source models locally \cite{wolf2020transformers}. 

\subsection{Hyper Parameters}
We use the 8-bit quantized AdamW optimizer~\cite{dettmers20218,kingma2017adam} for fine-tuning the models. We use the same fine-tuning strategy and hyperparameters for all the models that are trained locally. For inference, the maximum output tokens for the non-reasoning models is set to 256. For fine-tuning we use a learning rate of $2 \times 10^{-4}$ with a maximum step size of 60 and weight decay with a linear scheduler for all the models. We use gradient accumulation with steps $4$ and only fine-tune for 1 epoch since further training did not meaningfully improve the validation loss. To maximize GPU memory utilization with respect to model size, we make use of Flash attention \cite{dao2022flashattention} and quantized low rank adaptors \cite{dettmers2024qlora}. The adaptors are applied as Q, K, V, O, Gate, Up and Down projectors with hidden dimension size of 128 for all small and medium models and 64 for large models (the latter only because 128 could not fit in memory on the RTX 6000 Ada). 

For the reasoning Qwen models in Table~\ref{table:reasoning-results}, we set the maximum output tokens to 8192, and for o3-mini this is set to 15000.   

\subsection{Data Statistics}
\begin{figure*}[!h]
    \centering 
\begin{subfigure}{0.25\textwidth}
  \centerline{\includegraphics[width=0.55\linewidth]{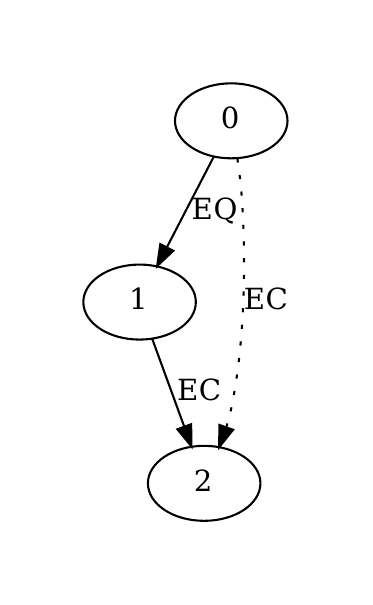}}
  \caption{$k=2, b=1$}
  \label{fig:1}
\end{subfigure}\hfil 
\begin{subfigure}{0.25\textwidth}
  \includegraphics[width=0.8\linewidth]{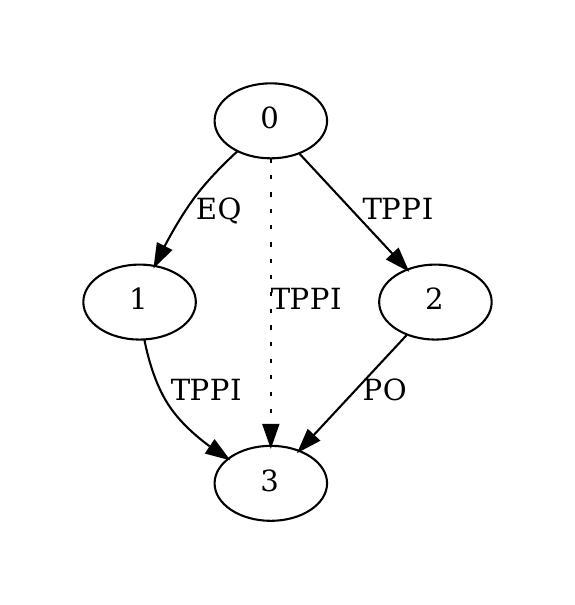}
  \caption{$k=2, b=2$}
  \label{fig:2}
\end{subfigure}\hfil 
\begin{subfigure}{0.25\textwidth}
  \includegraphics[width=\linewidth]{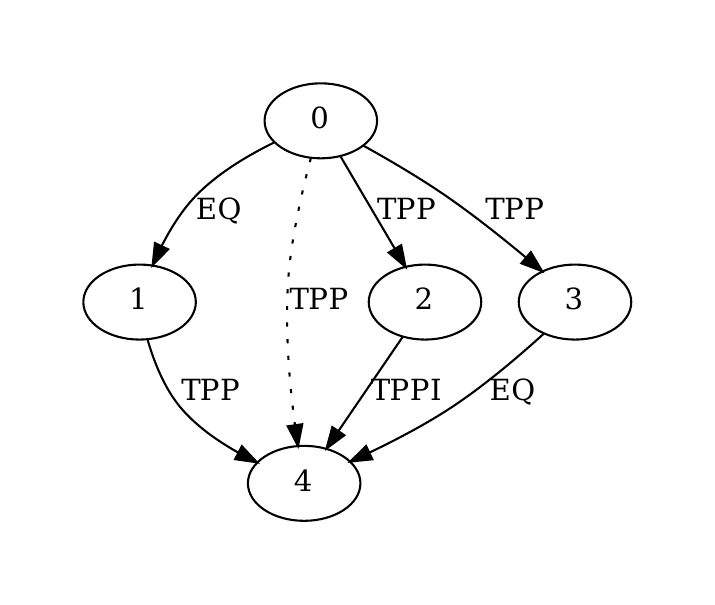}
  \caption{$k=2, b=3$}
  \label{fig:3}
\end{subfigure}
\begin{subfigure}{0.25\textwidth}
  \includegraphics[width=0.7\linewidth]{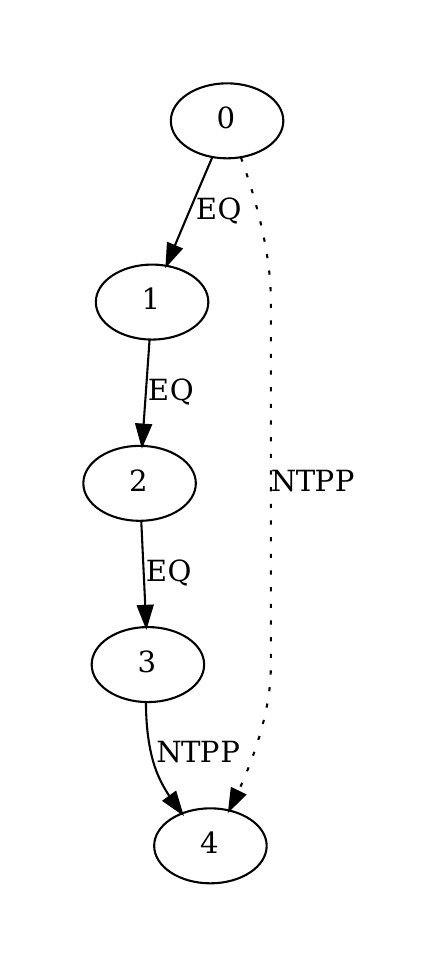}
  \caption{$k=4, b=1$}
\end{subfigure}\hfil 
\begin{subfigure}{0.25\textwidth}
  \includegraphics[width=0.9\linewidth]{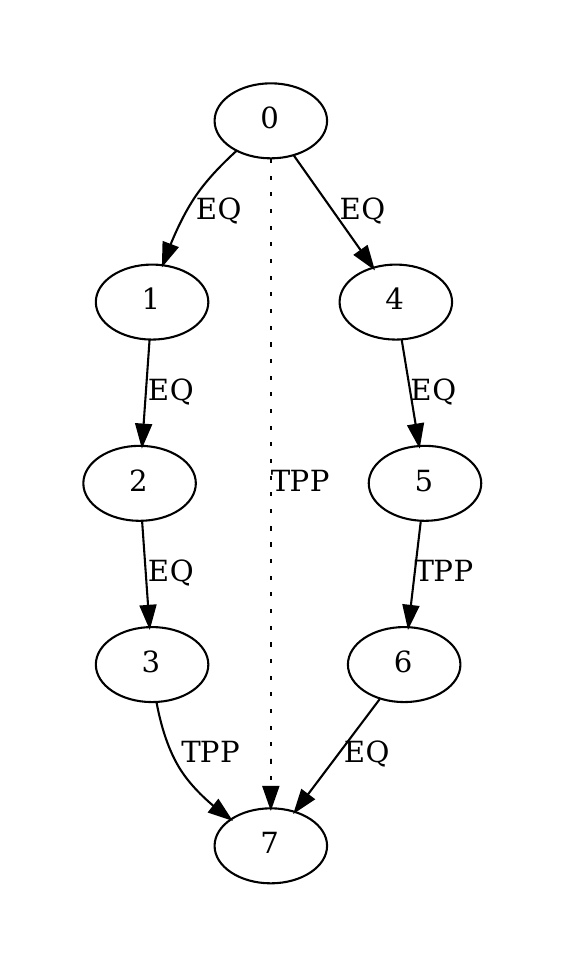}
  \caption{$k=4, b=2$}
\end{subfigure}\hfil 
\begin{subfigure}{0.25\textwidth}
  \includegraphics[width=1.\linewidth]{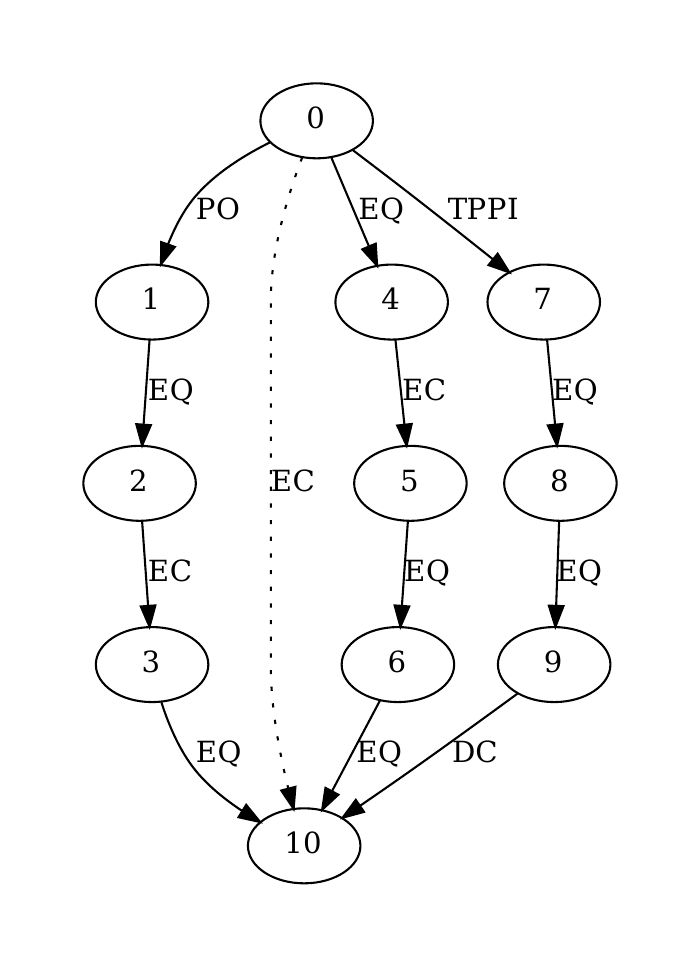}
  \caption{$k=4, b=3$}
\end{subfigure}
\begin{subfigure}{0.25\textwidth}
\hspace{-6ex}
\rightline{\includegraphics[width=1.2\linewidth]{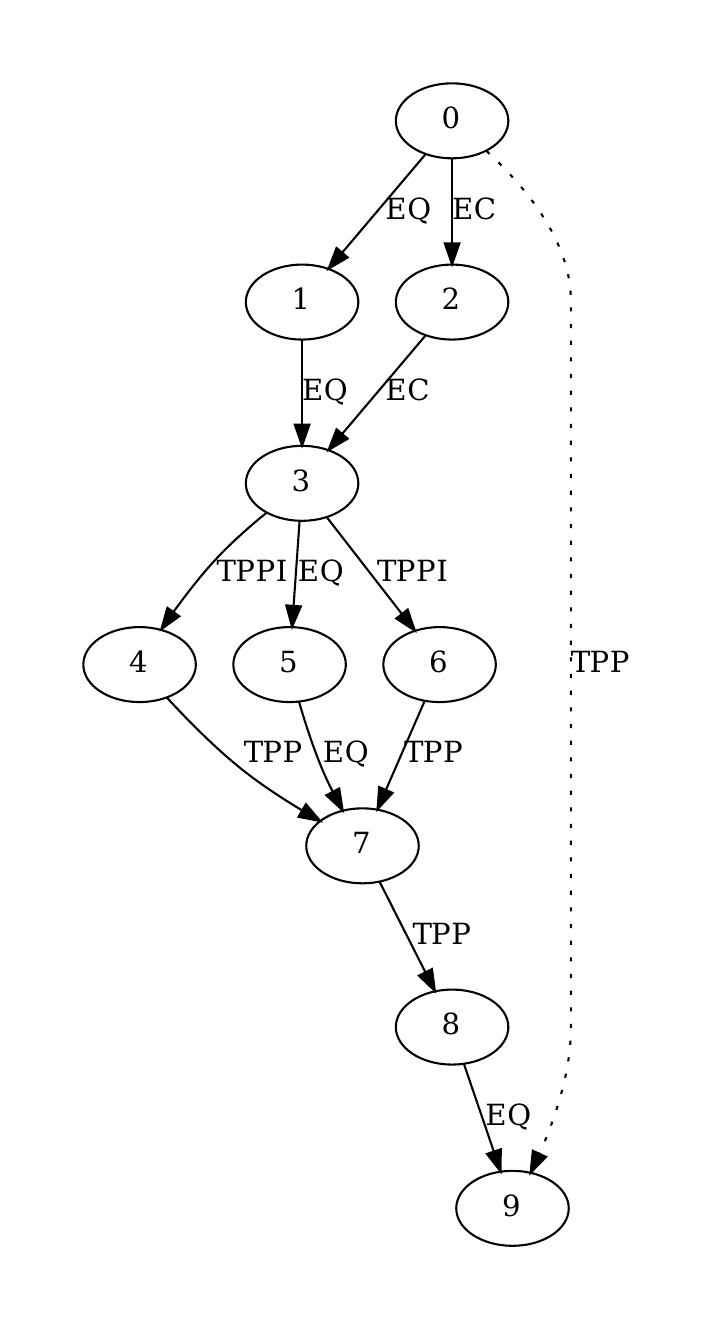}}
  \caption{$k=6, b=1$}
\end{subfigure}\hfil 
\begin{subfigure}{0.25\textwidth}
  \hspace{-4ex}
  \centerline{\includegraphics[width=2.7\linewidth]{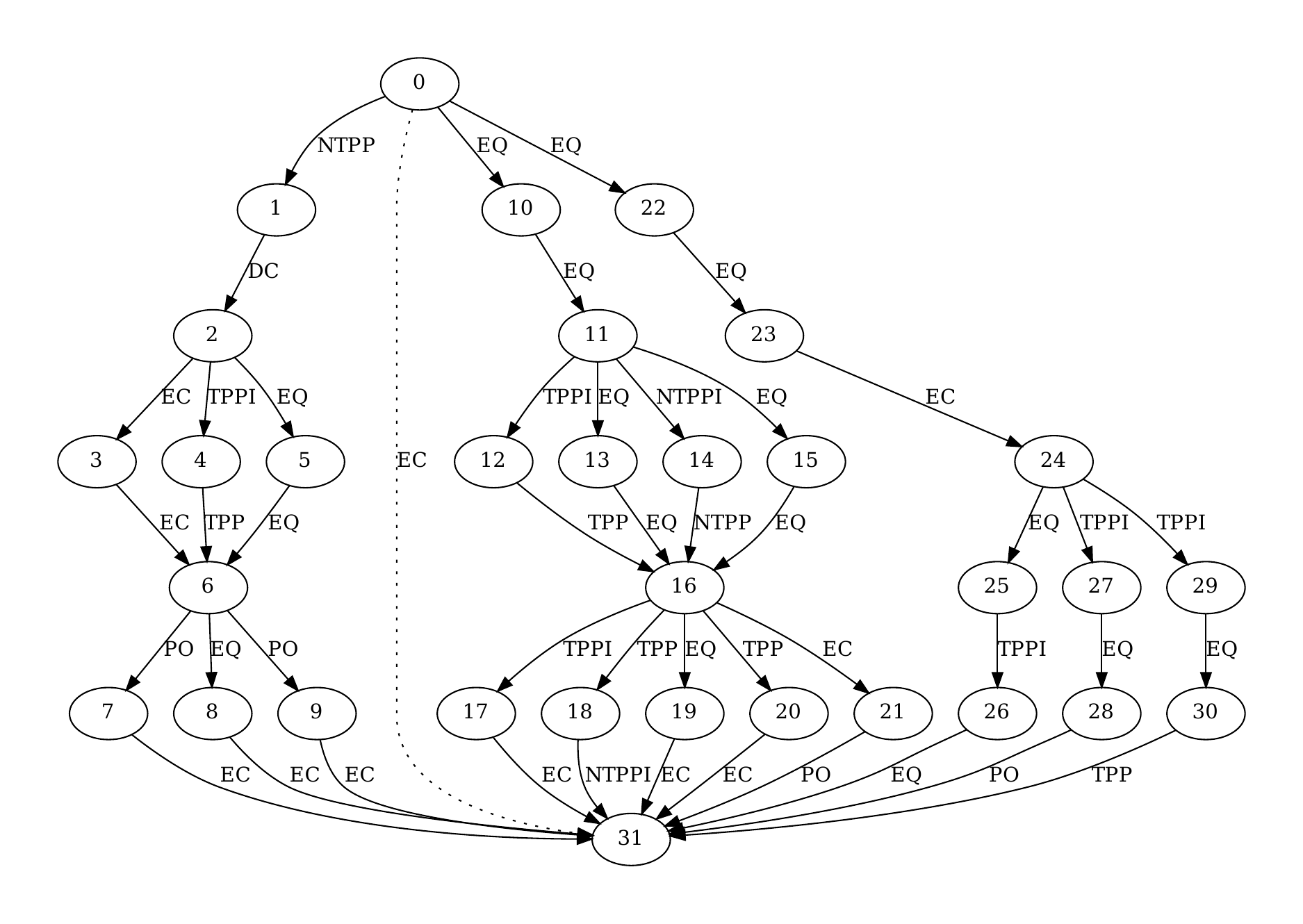}}
  \caption{$k=6, b=3$}
\end{subfigure}
\caption{Some graph instances for the RCC-8 dataset generated using the procedure described in~\cite{khalid2025systematic}. The target edge label between the source node and the tail node that needs to be predicted by the model is indicated by the dotted line.}
\label{fig:example-topo-rcc8}
\end{figure*}

The dataset statistics for the STaR benchmark for the training and test sets are summarized in the Table~\ref{table:rcc8-stats}. These are respectively subsampled for the experimental evaluations in the main text.
All random sampling is done with a global seed of 0 for reproducibility. Some example graphs generated via this procedure for the RCC-8 dataset are displayed in Figure~\ref{fig:example-topo-rcc8}.

\begin{table*}[t]
\footnotesize
\caption{Data statistics of the STaR reasoning datasets. These are respectively subsampled for the experimental valuations in the main text.}
\label{table:rcc8-stats}
\begin{center}
\begin{tabular}{lccccc}
\toprule
\textbf{Dataset} & \textbf{Training regime} &\textbf{No. of relations} &\textbf{\# Train} &\textbf{\# Test per config.} & \textbf{Test regime}\\
\midrule
RCC-8 & $b\in\{1,2,3\}, k\in\{2,3\}$ & 8 & 57,600 & 6,400 & $b\in\{1,\dots,4\}, k\in\{2,\dots,10\}$ \\
IA & $b\in\{1,2,3\}, k\in\{2,3\}$ & 13 & 93,400 & 9,300 & $b\in\{1,\dots,4\}, k\in\{2,\dots,10\}$ \\
\bottomrule
\end{tabular}
\end{center}
\end{table*}

\subsection{Prompts}
The prompts used for non-fine tuning experiments for RCC-8 are shown in Fig.~\ref{fig:rcc-8 prompt} with mutatis mutandis changes for IA  and for IA for the instruction-tuning setting in Fig.~\ref{fig:interval-algebra-prompt} with similar changes for RCC-8. We experimented with textual graph labels as opposed to integers in the prompt and the requested output format but found the accuracy and the adherence of the small models to be extremely poor in this setting with very low accuracies.
\begin{figure*}
    \centering
    \caption{The given prompt is for the inference RCC-8 dataset, while the Interval prompt for inference has a similar structure but different base elements and composition table.}
    \label{fig:rcc-8 prompt}
    \begin{prompt}{RCC8 Inference Prompt}{}
\begin{verbatim}

System: You are a helpful assistant. Just answer the question as a single integer.
\end{verbatim}
\noindent\rule{\linewidth}{0.5pt}
\begin{verbatim}
        
User: You are a qualitative spatial and temporal reasoning expert specializing in
RCC‑8 

The following are the base elements of RCC-8:

    DC = 1
    EC = 2
    PO = 4
    TPP = 8
    NTPP = 16
    TPPI = 32
    NTPPI = 64
    EQ = 128

The following is the composition table of RCC-8 as a JSON dictionary:
{(1, 1): [], (1, 2): [1, 2, 4, 8, 16], ..., (128, 64): [64], (128, 128):
[128]} 

Now the question is: Given a consistent graph with edges comprising the 8 
base relations, predict the label of the target edge. More specifically,
Given a data row delimited by a comma with the following columns:
`graph_edge_index`, `edge_labels`, `query_edge`, predict the label of the 
`query_edge` as one of the 8 base relations as a power of 2 as defined above. 
\end{verbatim}
\noindent\rule{\linewidth}{0.5pt}
       
\begin{verbatim}
(The optional few-shot examples:
Example 1: 
[(0, 1), (1, 2)], ['EQ', 'NTPPI'], (0, 2)
64

...

Example 5: 
[(0, 1), (1, 2), (2, 3)], ['EQ', 'EQ', 'EC'], (0, 3)
2
Examples end here.
)
\end{verbatim}
\noindent\rule{\linewidth}{0.5pt}
\begin{verbatim}
      
[(0, 1), (1, 4), (0, 2), (2, 4), (0, 3), (3, 4)], 
['EQ', 'NTPPI', 'EQ', 'NTPPI', 'TPPI', 'NTPPI'], (0, 4)

\end{verbatim}

\end{prompt}
\end{figure*}

% ...


      
      



\begin{figure*}
    \centering
    \caption{The given prompt is for the finetuining interval dataset, while the RCC-8 prompt for finetuning has a similar structure but different base elements and composition table.}
    \label{fig:interval-algebra-prompt}
    \begin{prompt}{Interval Finetuning Prompt}{}
    \begin{verbatim}
Below is an instruction that describes a task, paired with an input that provides 
further context. Write a response that appropriately completes the request.

### Instruction:
You are a qualitative spatial and temporal reasoning expert specializing in 
Interval Algebra.

The following are the base elements of Interval Algebra:

    '=': 1
    '<': 2 
    '>': 4 
    'd': 8
    'di': 16 
    'o': 32 
    'oi': 64 
    'm': 128 
    'mi': 256 
    's': 512
    'si': 1028 
    'f': 2048 
    'fi': 4096

The following is the composition table of RCC-8 as a JSON dictionary:
(eq, eq): [eq], (eq, lt): [lt],, ..., (fi, gt): [gt, oi, di, mi, si]}

Now the question is: Given a consistent graph with edges comprising the 8 
base relations, predict the label of the target edge. More specifically,
Given a data row delimited by a comma with the following columns:
`graph_edge_index`, `edge_labels`, `query_edge`, predict the label of the 
`query_edge` as one of the 8 base relations as a power of 2 as defined above. 


### Input:
[(0, 1), (1, 4), (0, 2), (2, 4), (0, 3), (3, 4)],
['m', '>', 'di', 'fi', 'di', 'oi'], (0, 4)

### Response:
16
        \end{verbatim}

    \end{prompt}
\end{figure*}

\section{Additional Analysis}
\subsection{Fine-grained breakdowns}
\begin{figure*}
    \centering
    \includegraphics[width=0.6\linewidth]{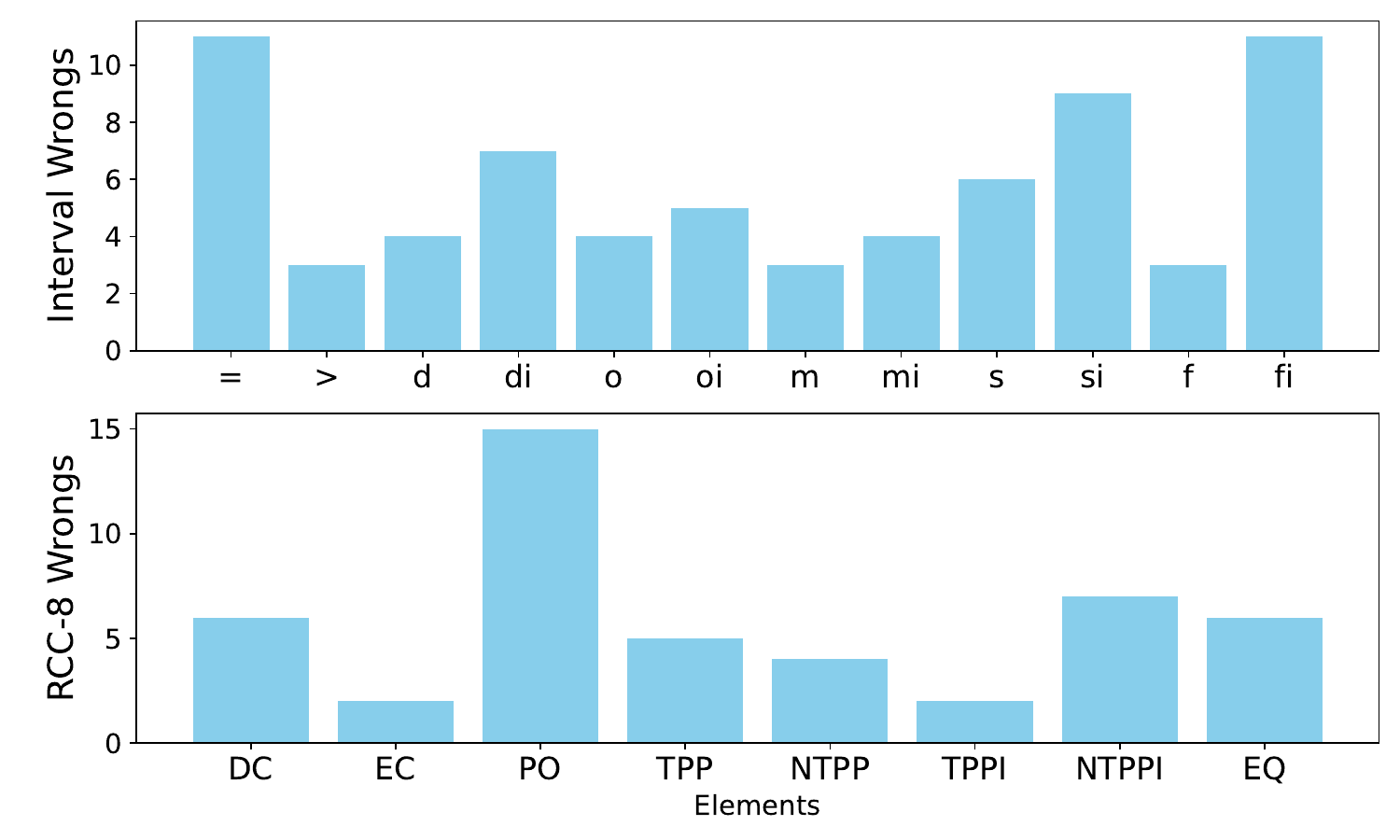}
    \caption{Non-responses from o3 where it took longer than the maximum allotted number of tokens. Certain classes are overrrepresented and for IA coincide with those that can easily predicted by leveraging heuristics based on dataset construction constraints.}
    \label{fig:wrongs-o3-mini}
\end{figure*}

Conducting a fine-grained classification level analysis of o3-mini for the instances where it thought for longer than 15000 tokens and responded with nothing over all the reasoning datasets is shown in figure~\ref{fig:wrongs-o3-mini}. We find that o3-mini took unexpectedly longer for the trivial relations such as $=$, and for and $\mathsf{fi}$ for IA and $\mathsf{po}$ for RCC-8. 
\subsection{Shallow ACA simulation}\label{app:aca-analysis}
\begin{figure*}
    \centering
    \includegraphics[width=0.5\linewidth]{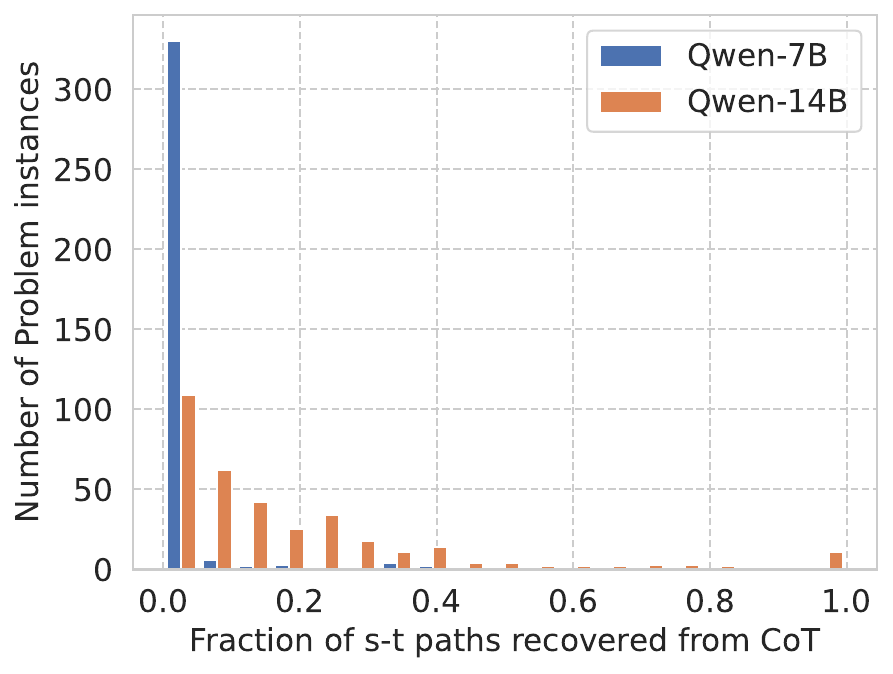}
    \caption{Fraction of source-to-tail paths recovered from the model's CoT for IA.}
    \label{fig:path-recovery-on-rcc8}
\end{figure*}

\begin{figure*}
    \centering
    \includegraphics[width=0.5\linewidth]{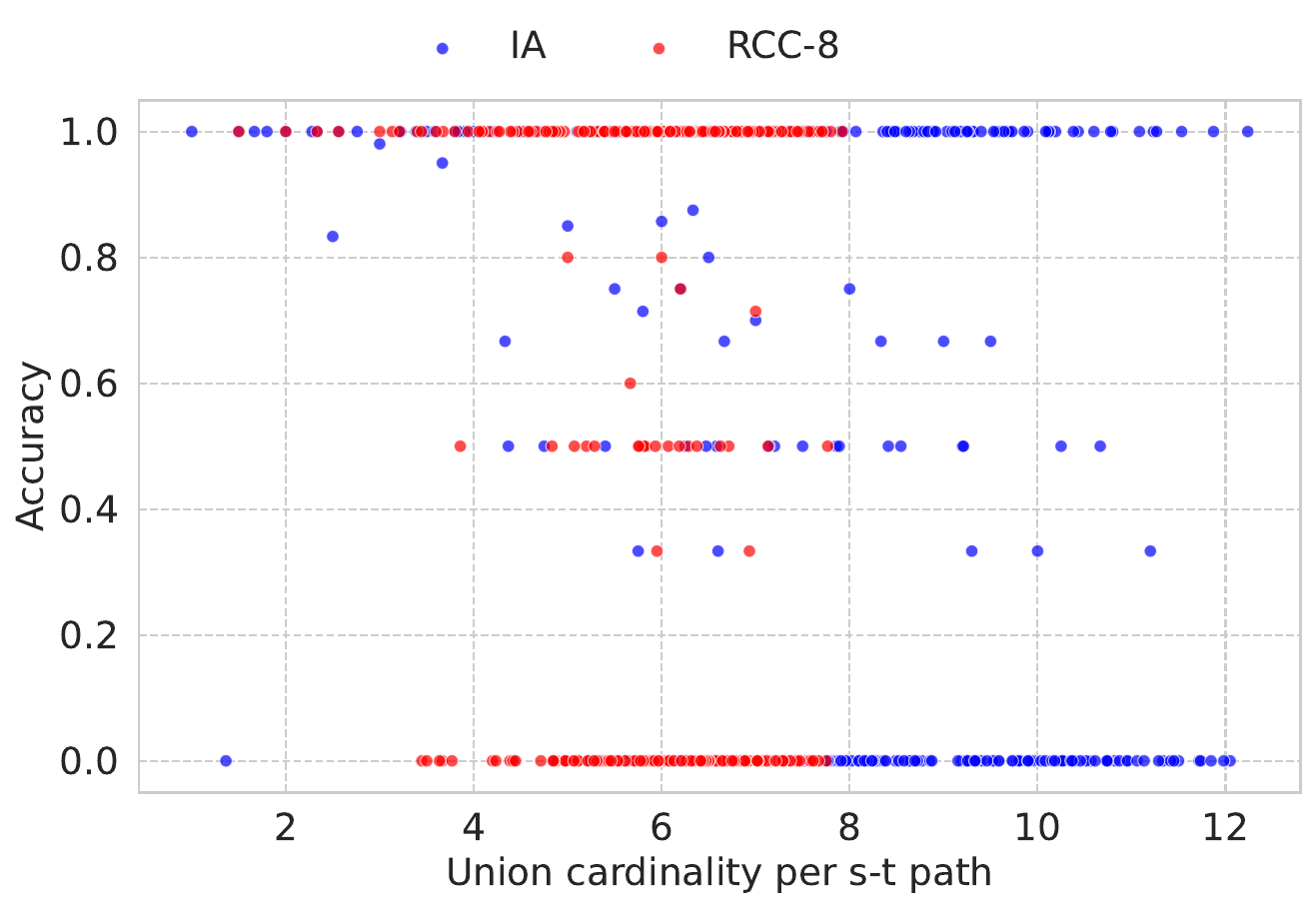}
    \caption{o3-mini performance as a function of union operations required per problem instance.}
    \label{fig:union-op-trend}
\end{figure*}

We show the fraction of s-t paths recovered from the CoT for the RCC-8 dataset in Figure~\ref{fig:path-recovery-on-rcc8}. The variation of o3-mini's performance with respect to the number of union operations is shown in Figure~\ref{fig:union-op-trend}. We measure union indirectly by computing the average over s-t paths of the cardinality (size) of the final set of multiple possible relations after all the chained relational compositions per path for a single problem instance.

\clearpage
\onecolumn
\section{An example CoT for an RCC-8 problem instance from R1 distilled Qwen-14B}
\label{app:cot-example}
\begin{cotbox}
\begin{lstlisting}

Okay, so I'm trying to figure out the label for the query edge (0, 20) in the given graph. The graph has several edges with known labels, and I need to use the RCC-8 composition table to determine the possible label for the target edge.

First, let me list out the edges and their labels as provided:

1. (0,1) - EQ (128)
2. (1,4) - EQ (128)
3. (0,2) - NTPPI (64)
4. (2,4) - NTPP (16)
5. (0,3) - TPPI (32)
6. (3,4) - TPP (8)
7. (4,5) - EC (4)
8. (5,10) - EC (4)
9. (4,6) - EC (4)
10. (6,10) - TPP (8)
11. (4,7) - TPPI (32)
12. (7,10) - TPP (8)
13. (4,8) - EQ (128)
14. (8,10) - EQ (128)
15. (4,9) - NTPPI (64)
16. (9,10) - NTPP (16)
17. (10,11) - DC (1)
18. (11,16) - PO (2)
19. (10,12) - DC (1)
20. (12,16) - PO (2)
21. (10,13) - TPP (8)
22. (13,16) - EQ (128)
23. (10,14) - EQ (128)
24. (14,16) - DC (1)
25. (10,15) - NTPP (16)
26. (15,16) - NTPPI (64)
27. (16,17) - DC (1)
28. (17,18) - EQ (128)
29. (18,20) - EQ (128)
30. (17,19) - EQ (128)
31. (19,20) - EQ (128)

I need to find the label for edge (0,20). Let's see how 0 and 20 are connected.

Looking at the graph, node 0 is connected to 1, 2, 3, and so on. Node 20 is connected to 18 and 19, which are both connected to 17. Node 17 is connected to 16 and 18. Node 16 is connected to 11, 12, 13, 14, 15, and 17.

So, the path from 0 to 20 could go through several nodes. Let's try to find the shortest path or the most direct path.

One possible path is 0 -> 1 -> 4 -> 5 -> 10 -> 11 -> 16 -> 17 -> 18 -> 20.

Another path is 0 -> 2 -> 4 -> 5 -> 10 -> 11 -> 16 -> 17 -> 18 -> 20.

Alternatively, 0 -> 3 -> 4 -> 5 -> 10 -> 11 -> 16 -> 17 -> 18 -> 20.

I think the key is to find the composition of relations along a path from 0 to 20 and see what possible labels can result from the composition.

Let's take the path 0 -> 1 -> 4 -> 5 -> 10 -> 11 -> 16 -> 17 -> 18 -> 20.

Each edge has a label, and we can compose these labels step by step using the composition table.

Starting with the edge (0,1): EQ (128)
Then (1,4): EQ (128)
So, EQ composed with EQ. Looking at the composition table, (128,128) isn't directly given, but I can check the entries. Wait, the composition table is given as a dictionary where each key is a tuple (a, b), and the value is a list of possible results.

Wait, no, actually, the composition table is for the base relations. Each entry is (x, y) mapping to a list of possible z where x R y and y S z implies x T z.

But maybe I'm overcomplicating. Alternatively, since each edge's label is a base relation, we can compose them step by step.

Let me think of the composition as follows: each step, we take the current relation and compose it with the next edge's relation.

Starting from 0 to 1: EQ (128)
Then 1 to 4: EQ (128)
So, EQ composed with EQ. Looking at the composition table, for (EQ, EQ), which is (128,128), but the table is structured as (a, b) where a and b are the labels. Wait, no, the composition table is for the base elements, not the powers. Wait, the base elements are DC=1, EC=2, PO=4, etc. So, the labels are powers of 2, and the composition table is for the base relations, not the labels directly.

Wait, perhaps I should convert the labels to their base relation numbers first.

EQ is 128, which is NTPPI (since NTPPI is 64, wait no, the base elements are:

DC = 1

EC = 2

PO = 4

TPP = 8

NTPP = 16

TPPI = 32

NTPPI = 64

EQ = 128

So, each label is one of these numbers. The composition table is given as a dictionary where each key is (a, b), and the value is the list of possible c such that a R b and b S c implies a T c.

But I'm not sure if I should compose the relations step by step or consider the entire path.

Alternatively, perhaps I can find the relations between 0 and 20 through other nodes.

Looking at the graph, 0 is connected to 1, 2, 3, and others. Let's see:

From 0 to 2: NTPPI (64)
From 2 to 4: NTPP (16)
From 4 to 5: EC (4)
From 5 to 10: EC (4)
From 10 to 11: DC (1)
From 11 to 16: PO (2)
From 16 to 17: DC (1)
From 17 to 18: EQ (128)
From 18 to 20: EQ (128)

So, the path from 0 to 20 via 2 is:

0-2: 64 (NTPPI)
2-4: 16 (NTPP)
4-5:4 (EC)
5-10:4 (EC)
10-11:1 (DC)
11-16:2 (PO)
16-17:1 (DC)
17-18:128 (EQ)
18-20:128 (EQ)

Now, let's compose these step by step.

Start with 64 (NTPPI).

Then, compose with 16 (NTPP). So, (64,16). Looking at the composition table, I need to find the entry for (NTPPI, NTPP). Wait, the table is given as a dictionary with tuples of the label numbers. So, NTPPI is 64, NTPP is 16.

Looking at the composition table, the key (64,16) would be (64,16). But in the given composition table, I don't see an entry for (64,16). Wait, let me check the table:

Looking through the composition table entries, I see:

(64, 16): Not directly listed, but perhaps I can find it by checking the entries.

Wait, the composition table is a dictionary where each key is (a, b), and the value is the list of possible c's.

Looking through the provided composition table, I don't see an entry for (64,16). Hmm, that might be an issue. Alternatively, perhaps I should consider that the composition of NTPPI (64) and NTPP (16) would result in some relation.

Wait, maybe I should look for the entry (64,16). Let me scan the table:

Looking at the list, I see entries like (64,16): not present. Wait, perhaps I should check all the entries.

Wait, the composition table is provided as a JSON dictionary, but in the user's message, it's a long list of tuples. Let me check if (64,16) is present.

Looking through the list:

Looking for (64,16). Scanning the list:

I see entries like (64,16) is not directly listed, but perhaps I can infer it from other entries.

Alternatively, perhaps I should think in terms of the base relations and their composition.

Wait, maybe I'm approaching this wrong. Let me try to compose the relations step by step.

Starting with 0-2: 64 (NTPPI)

Then 2-4:16 (NTPP)

So, NTPPI composed with NTPP. Let's look at the composition table for (64,16). Wait, the table doesn't have this entry. Hmm.

Alternatively, perhaps I should consider that NTPPI is 64, and NTPP is 16. The composition of NTPPI and NTPP would be the result of 64 R 16, where R is the relation between 64 and 16.

Wait, but in the composition table, each entry is (x, y) mapping to possible z's. So, to find what 64 R 16 can result in, I need to see if (64,16) is a key in the composition table.

Looking through the provided composition table, I don't see (64,16) as a key. The closest is (64,16) is not present. Hmm, that's a problem.

Wait, perhaps I should consider that the composition of NTPPI (64) and NTPP (16) would result in some relation. Let me think about the possible results.

Wait, perhaps I should look for the entry (64,16) in the composition table. Since it's not present, maybe the composition is empty, which would mean that the path 0-2-4 is not possible. But that can't be right because the graph is consistent.

Alternatively, maybe I'm misunderstanding the composition table. Let me check the table again.

Looking at the composition table, the entries are:

For example, (1,1): [], (1,2): [1,2,4,8,16], etc.

So, each entry is (a, b) mapping to a list of possible c's where a R b implies a T c via b.

So, for (64,16), I need to find if there's an entry in the composition table for (64,16). If not, perhaps it's not directly composable, but maybe through other steps.

Alternatively, perhaps I should consider that the composition of 64 (NTPPI) and 16 (NTPP) would result in some relation, but since it's not in the table, maybe the only possible result is 64 (NTPPI) composed with 16 (NTPP) would be NTPPI, but I'm not sure.

Alternatively, perhaps I should take a different path.

Another path from 0 to 20 is through node 3.

0-3:32 (TPPI)
3-4:8 (TPP)
4-5:4 (EC)
5-10:4 (EC)
10-11:1 (DC)
11-16:2 (PO)
16-17:1 (DC)
17-18:128 (EQ)
18-20:128 (EQ)

So, composing these:

32 (TPPI) composed with 8 (TPP). Let's see (32,8) in the composition table.

Looking at the composition table, I see (32,8) is not directly listed, but perhaps I can find it.

Wait, looking through the entries, I see (32,8) is not present. Hmm.

Alternatively, perhaps I should look for the entry (32,8). Since it's not present, maybe the composition is not possible, but that can't be right.

Alternatively, perhaps I should consider that the composition of TPPI (32) and TPP (8) would result in something.

Wait, in the composition table, I see (32,8) is not present, but perhaps I can infer it from other entries.

Alternatively, maybe I should take another path.

Another path: 0-1:128 (EQ)
1-4:128 (EQ)
4-5:4 (EC)
5-10:4 (EC)
10-11:1 (DC)
11-16:2 (PO)
16-17:1 (DC)
17-18:128 (EQ)
18-20:128 (EQ)

So, starting with 0-1:128 (EQ)
Then 1-4:128 (EQ)
So, EQ composed with EQ. Looking at the composition table, (128,128) is not directly listed, but perhaps I can find the result.

Wait, the composition table doesn't have (128,128) as a key. Hmm.

Alternatively, perhaps I should consider that EQ composed with EQ is EQ, so 128.

Then, 4-5:4 (EC)
So, EQ (128) composed with EC (4). Let's look for (128,4) in the composition table.

Looking through the entries, I see (128,4) is not present. Hmm.

Alternatively, perhaps I should think that EQ composed with EC would result in EC, but I'm not sure.

Alternatively, maybe I should consider that the composition of EQ and EC is EC, so 4.

But I'm not sure. This is getting complicated.

Alternatively, perhaps I should consider that the query edge (0,20) is connected through multiple paths, and I need to find a relation that is consistent across all possible paths.

But this might take a lot of time. Alternatively, perhaps I can look for the relations that are common in the paths.

Wait, another approach: since the graph is consistent, the label for (0,20) must be such that all paths from 0 to 20 agree on the same label.

Alternatively, perhaps I can find the relations along the path and see what the composition would be.

Let me take the path 0-1-4-5-10-11-16-17-18-20.

The relations are:

0-1:128 (EQ)
1-4:128 (EQ)
4-5:4 (EC)
5-10:4 (EC)
10-11:1 (DC)
11-16:2 (PO)
16-17:1 (DC)
17-18:128 (EQ)
18-20:128 (EQ)

Now, let's compose these step by step.

Start with 128 (EQ).

Then, compose with 128 (EQ). Let's see what (128,128) gives. Since it's not in the table, perhaps the result is 128.

Then, compose with 4 (EC). So, (128,4). Looking for (128,4) in the composition table. Not present. Hmm.

Alternatively, perhaps the composition of EQ (128) and EC (4) would result in EC (4).

Then, compose with 4 (EC). So, 4 composed with 4 is (4,4). Looking at the composition table, (4,4) is not present, but perhaps the result is 4.

Then, compose with 1 (DC). So, (4,1). Looking at the composition table, (4,1) is present and the result is [1,2,4,32,64].

So, possible results are 1,2,4,32,64.

Then, compose with 2 (PO). So, for each possible result from the previous step, we need to compose with 2.

Wait, this is getting too complicated. Maybe I should try a different approach.

Alternatively, perhaps the query edge (0,20) is connected through multiple paths, and the only consistent label is one that is possible in all paths.

Looking at the possible paths, the relations along the way include EQ, EC, DC, PO, etc.

But I'm not making progress. Maybe I should look for the relations that are common in the paths.

Alternatively, perhaps the label for (0,20) is EQ (128), as many edges along the path are EQ.

But I'm not sure. Alternatively, perhaps I should consider that the composition of all the relations along the path would result in EQ.

Wait, another idea: the path from 0 to 20 through 17-18-20 has EQ relations. So, maybe the final composition would be EQ.

Alternatively, perhaps the label is EQ (128).

But I'm not certain. Alternatively, maybe it's DC (1), but that seems less likely.

Wait, considering the path 0-1-4-5-10-11-16-17-18-20, the last few steps are DC, PO, DC, EQ, EQ.

So, composing DC (1) with PO (2) gives (1,2) which from the composition table is [1,2,4,8,16].

Then, compose with DC (1): So, for each possible result from (1,2), which is [1,2,4,8,16], we compose with 1.

Looking at (x,1) where x is in [1,2,4,8,16], the results would be:

From the composition table:

(1,1): []
(2,1): [1,2,4,32,64]
(4,1): [1,2,4,32,64]
(8,1): [1]
(16,1): [1]

So, for each x in [1,2,4,8,16], (x,1) gives:

1: []
2: [1,2,4,32,64]
4: [1,2,4,32,64]
8: [1]
16: [1]

So, the possible results are 1,2,4,32,64.

Then, compose with EQ (128) at 17-18: So, for each possible result from the previous step, we compose with 128.

Wait, but 17-18 is 128 (EQ). So, the previous step's result is composed with 128.

But I'm getting lost here.

Alternatively, perhaps the final composition would be EQ (128), as the last step is EQ.

But I'm not sure. Alternatively, perhaps the label is EQ (128).

But I'm not certain. Maybe I should look for another approach.

Wait, considering that node 20 is connected to 18 and 19, which are both connected to 17. 17 is connected to 16 and 18. 16 is connected to 11,12,13,14,15,17.

Looking at the edges from 16 to 17: DC (1).

From 17 to 18: EQ (128)
From 17 to 19: EQ (128)
From 18 to 20: EQ (128)
From 19 to 20: EQ (128)

So, the path from 16 to 20 is through 17-18 or 17-19, both with EQ.

So, the edge from 16 to 20 is through 17, which is DC (1) to 17, then EQ (128) to 18 or 19, and then EQ (128) to 20.

So, the composition from 16 to 20 would be DC (1) composed with EQ (128) composed with EQ (128).

So, first, (1,128): Looking at the composition table, (1,128) is not present, but perhaps the result is 1, as DC composed with EQ might be EQ.

Wait, but I'm not sure. Alternatively, perhaps the composition of DC (1) and EQ (128) would result in EQ (128), since DC is "disjoint" and EQ is "equal", which might not make sense, but perhaps it's possible.

Alternatively, perhaps the composition is 128.

Then, composing with another EQ (128) would give 128.

So, the result from 16 to 20 would be 128 (EQ).

Then, going back to 10-11: DC (1)
11-16: PO (2)
So, DC (1) composed with PO (2). Looking at (1,2), which is [1,2,4,8,16].

Then, compose with the result from 16 to 20, which is 128.

Wait, but 16 to 20 is 128, so the path from 10 to 20 would be DC (1) composed with PO (2) composed with 128.

So, first, (1,2) gives [1,2,4,8,16].

Then, each of these composed with 128.

For each x in [1,2,4,8,16], (x,128) would be:

Looking at the composition table:

(1,128): Not present, but perhaps the result is 128.
(2,128): [2]
(4,128): [4]
(8,128): [8]
(16,128): [16]

So, the possible results would be 128, 2, 4, 8, 16.

Then, the next step is from 10 to 20, which would be the composition of DC (1), PO (2), and 128.

But I'm getting stuck here.

Alternatively, perhaps I should consider that the query edge (0,20) is connected through multiple paths, and the only consistent label is EQ (128), as many edges along the path are EQ.

Therefore, I think the label for the query edge (0,20) is EQ, which is 128.
</think>

The query edge (0,20) is determined by composing the relations along the path from 0 to 20. After considering multiple paths and their compositions, the most consistent result is EQ (128).


```json
{"label": 128}
```
\end{lstlisting}
\end{cotbox}
\twocolumn
\end{document}